\documentclass[11pt]{article}

\usepackage[final]{acl}

\usepackage{times}
\usepackage{latexsym}

\usepackage[T1]{fontenc}

\usepackage[utf8]{inputenc}

\usepackage{microtype}

\usepackage{inconsolata}

\usepackage{graphicx}
\usepackage{amsmath}
\usepackage{amssymb}
\usepackage{booktabs}
\usepackage{tabularx}
\usepackage{mathtools}
\usepackage{xcolor}
\usepackage{stfloats}
\usepackage{url}
\usepackage[normalem]{ulem}

\newcommand{\method}[0]{SP-EWC}
\newcommand{\longmethod}[0]{Subspace-Projected Elastic Weight Consolidation}

\title{Protecting Safety during Reasoning Training \\ with Low-Rank Fisher Regularisation}

\author{
    Neel Rajani\textsuperscript{1} \and Ivan Titov\textsuperscript{1,2} \\
    \textsuperscript{1}ILCC, University of Edinburgh\\ \textsuperscript{2}ILLC, University of Amsterdam\\
    \texttt{Neel.Rajani(at)ed.ac.uk}
    }

\begin{document}
\maketitle
\begin{abstract}

Large language models are trained to have safety guardrails, but these can be compromised while fine-tuning for new capabilities, such as reasoning. One way to mitigate this degradation is to modify the fine-tuning objective to preserve safety behaviour, but doing so effectively and efficiently remains challenging. We revisit Elastic Weight Consolidation (EWC), which penalises parameter updates based on sensitivity estimates from the diagonal of the Fisher information matrix (FIM). While EWC provides some protection, the diagonal approximation ignores interactions between parameters. We address this limitation by exploiting recent findings that safety-relevant behaviour is located in a low-rank subspace within models. Our method, \longmethod~(\method), projects gradients onto this subspace to capture non-diagonal curvature efficiently. We investigate three different ways of identifying the safety subspace: through safety fine-tuning updates (LoRA), safety-related activations (ActSVD), and contrastive representations of safety and utility data (cPCA). We find that this choice substantially affects performance, with \method~using cPCA comparing favourably against existing methods by better maintaining safety guardrails without introducing excessive over-refusal.\footnote{We commit to releasing our code and models following peer-review.}

\end{abstract}

\section{Introduction}
\label{sec:Introduction}

\begin{figure}[h]
    \begin{center}
    \includegraphics[width=1\linewidth]{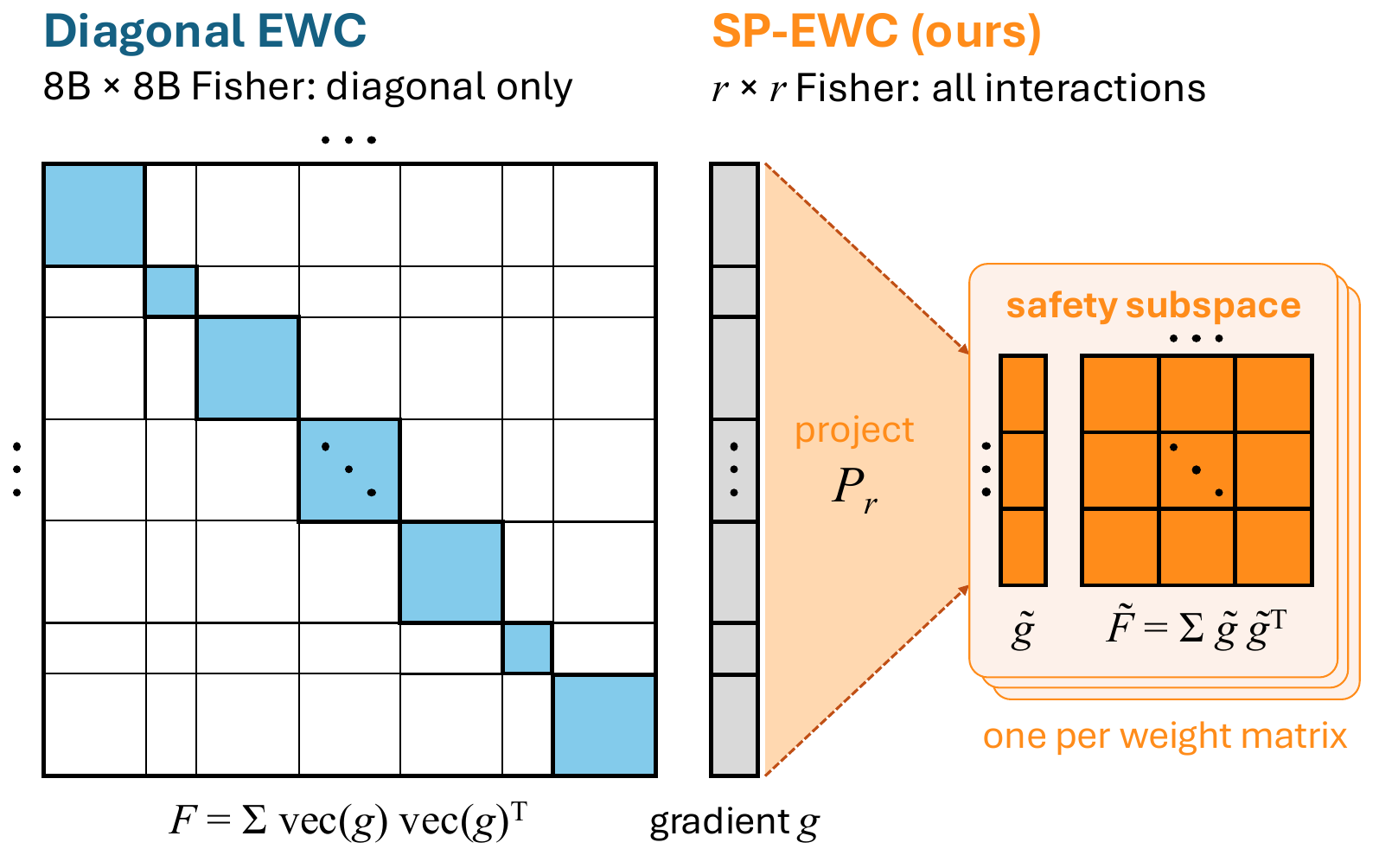}
    \end{center}
    \caption{The prohibitive size of the full Fisher information matrix  forces EWC to use a diagonal approximation, capturing only parameter-wise sensitivity. We instead project gradients onto the safety subspace, capturing interactions between safety-critical directions.}
    \label{fig:Intro figure}
\end{figure}

While the capabilities of large language models (LLMs) continue to advance dramatically, questions on how to ensure that more powerful models do not enable greater harms in society become more pressing. Specifically, as unprecedented amounts of compute are invested into complex post-training stages to train reasoning models \citep{semianalysisLongLive}, a key question is how to ensure that a model trained to be safe\footnote{Note that the term "safe" is contentious in this context, which we discuss further in Section \ref{sec:Locating the safety subspace correctly is key for protecting guardrails}.} does not lose its guardrails after further training. There is a growing body of work showing that even benign fine-tuning can compromise the safety of LLMs \citep{qi2024finetuning, zhan-etal-2024-removing, yang2024shadow}, and that this is also pronounced when teaching models to use reasoning \citep{jiang2025safechainsafetylanguagemodels, zhou2025hiddenriskslargereasoning, li2025finding, yong2026selfjailbreakinglanguagemodelsreason}. We reproduce this finding: training instruction-tuned models on $\sim$1B tokens of reasoning traces via supervised fine-tuning reduces their performance on jailbreak-resistance benchmarks by up to 75.4 percentage points (from an average on four benchmarks of 92.7\% down to 17.3\%, when using a higher learning rate). In response, recent research has investigated how to fine-tune models without degrading guardrails, such as mixing safety data into training \citep{huang2024Lisa, jiang2025safechainsafetylanguagemodels, fang2026grsapgenerativereplaysafety}, modifying gradient updates \citep{yi2026safegrad}, incorporating safety LoRAs \citep{hsu2024safe}, adding novel regularisers \citep{kim2025rethinking, peng-etal-2026-preventing}, or applying established continual learning (CL) methods \citep{alssum2025unforgottensafetypreservingsafety}, many of which also incorporate prior task data. However, we find that baselines  fail to protect jailbreak resistance, introduce new pathologies like over-refusal, or add considerable computational overhead.

We revisit a classical approach to continual learning,  Elastic Weight Consolidation (EWC; \citealp{elastic_weight_consolidation}). The core intuition behind EWC is to estimate how sensitive the model's performance on a prior task is to changes in its parameters. This estimate is then used to regularise training on a new task: changes to highly sensitive parameters incur a larger penalty, while less sensitive parameters are constrained less. This approach is promising for safety: after identifying parameters important for maintaining safe behaviour, their updates can be constrained during subsequent training. This can preserve refusal capabilities acquired during post-training while still allowing the model to learn new capabilities.

EWC incorporates parameter sensitivity via model curvature, normally by using a diagonal approximation of the Fisher information matrix (FIM). We find this to already mitigate some of the safety degradation but to not fully solve the problem. A key limitation is that the diagonal approximation captures the sensitivity of individual parameters but ignores interactions between them. Modelling the full covariance matrix across all parameter pairs instead would capture richer interactions between them, but becomes  infeasible even with small models. We propose to overcome this by exploiting growing evidence that safety-critical directions span a low-rank subspace within models \citep{arditi2024refusal, wei2024assessing, peng2024navigating, yang2025asft, perin2025lox}. We compute a Fisher within this safety subspace by projecting the gradients of each layer onto its safety directions. This yields a small $r \times r$ matrix which captures how these directions interact, unlike the mere parameter-wise estimates of the diagonal Fisher, as shown in Figure \ref{fig:Intro figure}. We then use this to penalise parameter drift during reasoning training with respect to the safety subspace, which we refer to as \textbf{\longmethod~(\method)}. We ask: 

\begin{itemize}
    \item[]\textit{Can we protect safety guardrails by exploiting the low dimensionality of their subspace and computing a non-diagonal approximation of the Fisher within it?} 
\end{itemize}

A key challenge, however, is how to identify the low-rank subspace associated with safety. Low-dimensionality can be operationalised in different ways. We investigate three approaches reflecting different notions of safety-relevant subspaces: directions associated with safety fine-tuning updates (LoRA; \citealp{hu2022lora}), directions prominent in safety-related activations after removing those associated with general utility (ActSVD; \citealp{wei2024assessing}), and directions that distinguish safety-related activations from those on benign inputs (contrastive PCA, cPCA; \citealp{Abid2018-iq}). We find that this choice is critical: LoRA and ActSVD do not consistently identify subspaces whose protection preserves safety, whereas cPCA performs more reliably across models. We summarise our key contributions as follows:

\begin{itemize}
    \item[] \textbf{Non-diagonal EWC in the safety subspace.} We exploit evidence that LLM safety is low-dimensional to make non-diagonal EWC practical, retaining interactions between safety-critical directions rather than only parameter-wise sensitivities.
    \item[] \textbf{Identifying safety subspaces.}   We investigate three approaches for
    identifying safety-relevant subspaces, including, to our knowledge, the first use of contrastive PCA~\cite{Abid2018-iq} as a training intervention.  We show that the choice of subspace is critical to the effectiveness of our method.
    \item[] \textbf{Maintaining guardrails without causing over-refusal.} Using cPCA to identify the safety subspace, our method achieves a better balance between task performance, safety preservation and avoidance of over-refusal than the considered alternatives.
\end{itemize}

\section{Related works}
\label{sec:Related_works}

\textbf{Guardrail removal from fine-tuning.} Safety guardrails are easily removed even by benign fine-tuning \citep{qi2024finetuning, zhan-etal-2024-removing, yang2024shadow, casper2026open}. This has motivated mitigations such as lowering the learning rate \citep{kim2025rethinking, lin2026sft}, mixing safety samples into the training dataset \citep{bianchi2024safetytuned, huang2024Lisa, fang2026grsapgenerativereplaysafety}, or extrapolating safety-critical directions \citep{perin2025lox}. They are complemented by findings that refusal is mediated by a single direction \citep{arditi2024refusal}, safety training mostly teaches the first few refusal tokens \citep{qi2025safety}, there is a 'safety basin' in weight space \citep{peng2024navigating}, and that there are sparse safety-critical weights that freezing cannot protect \citep{wei2024assessing}. We seek to exploit this low dimensionality to protect safety during fine-tuning.

\textbf{Elastic Weight Consolidation (EWC).} 
Following its introduction by \citet{elastic_weight_consolidation}, EWC became a popular baseline in continual learning \citep{Note2018Huszar, Reply2018Kirkpatrick, li2025fishers}. It has been applied to LLMs in contexts including regularised fine-tuning via world models \citep{Xiang2023EWC-world-models}, continued pre-training of Gemma 2 on Lithuanian data \citep{sliogeris2025fullparametercontinualpretraininggemma2}, and regularisation of shared and separate low-rank updates \citep{wei2025online, zheng2026revisiting}. Closest to our investigation is work by \citet{alssum2025unforgottensafetypreservingsafety}, who find standard EWC to not preserve safety guardrails very well. Our work aims to bridge this gap.

\textbf{Approximations of the Fisher information matrix and Hessian.} Approximating curvature information has been extensively studied in the context of second-order optimisation \citep{limitations2019, ven2025on}, with methods such as K-FAC \cite{pmlr-v37-martens15, martens2018kroneckerfactored}, Shampoo \cite{pmlr-v80-gupta18a}, and their variants \citep[e.g.,][]{peirson2022fishy}. Related approximations have also been explored for influence functions, including low-rank Hessian approximations \cite{Schioppa2022}. Such methods can be combined with EWC to capture interactions between parameters \citep{NEURIPS2018_f31b2046}, but introduce additional computational costs. Rather than relying on general-purpose curvature approximations, we exploit the low dimensionality of safety-relevant subspaces to capture interactions between safety-critical directions efficiently.

\section{Methods}
\label{sec:Methods}

\subsection{Problem setup}
\label{sec:Problem setup}
The general problem we tackle is that of a safe model losing its jailbreak resistance during fine-tuning. More specifically, consider an off-the-shelf model which has been trained for safety by its developers (such as \texttt{Llama-3.1-8B-it}), whose parameters we denote as $\theta^{ref}$. This model is then fine-tuned via supervised fine-tuning (SFT) on a dataset of $N$ prompts $x$ and responses $y$, $\mathcal{D}_{\text{SFT}} = \{(x_n, y_n) \}_{n=1}^N$, with the standard SFT loss computed on output tokens:

\begin{equation}
\label{eqn:SFT}
\mathcal{L}_{\text{SFT}}(\theta;\mathcal{D}_{\text{SFT}}) = -\mathbb{E}_{(x,y)\sim \mathcal{D}_{\text{SFT}}} [\log p_\theta (y|x)].
\end{equation}

As seen in prior work \citep{qi2024finetuning, zhan-etal-2024-removing, yang2024shadow} and replicated in our results below, this approach can inadvertently cause catastrophic forgetting of prior safety training. Our objective, therefore, is to find a way to maintain the safety of LLMs during fine-tuning, while also learning the task and minimising over-refusal. We assume access to a safety dataset $\mathcal{D}_{\text{Safety}}$, which consists of $M$ harmful prompts $x$ and safe responses $y$: $\mathcal{D}_{\text{Safety}} = \{ (x_j, y_j) \} _{j=1}^M$. Similarly, we assume availability of a utility dataset $\mathcal{D}_{\text{Util}}$.

\subsection{Elastic Weight Consolidation with diagonal Fisher information}
\label{sec:Elastic Weight Consolidation with diagonal Fisher information}

Our method extends Elastic Weight Consolidation (EWC). We refer readers to the original work by \citet{elastic_weight_consolidation} and further discussion in Appendix \ref{sec:Appendix - Further notes on computing Fisher Information for LLMs}.

EWC was proposed in order to maintain prior task performance while training on a new task. It does so by estimating the sensitivity of parameters with respect to the prior task (in our case, $\mathcal{D}_{\text{Safety}}$), and using this to regularise new updates to parameters. The sensitivity estimate is normally computed via the diagonal of the empirical Fisher information matrix (FIM), $D^{emp}$. While the method is general and the standard FIM is theoretically preferable, this is infeasible at the scale of modern models. At training step $t$, the movement of parameters of the model ($\theta^t$) away from the reference model ($\theta^{ref}$) is tracked as $\Delta_i = \theta_i^t - \theta_i^{ref}$ and weighted by the diagonal FIM and scaled by hyper-parameter$\lambda_\text{D}$, such that large changes to highly sensitive parameters are punished more:

\begin{equation}
    \label{eqn:EWC objective_background}
    \mathcal{L}_{\text{EWC}} = \mathcal{L}_{\text{SFT}} + \sum_i \frac{\lambda_\text{D}}{2} D_i^{emp} \Delta_i^2
\end{equation}

In our experiments, we move beyond the empirical FIM and instead estimate the true Fisher with model samples. Specifically, we construct $\mathcal{D}_{\text{Safety}}$ by sampling from $\theta^{ref}$, following recommendations by \citep{ven2025on}.

EWC became a key baseline in continual learning, but a key drawback is its reliance on the diagonal FIM. The full FIM is computationally intractable at LLM scale, so some form of approximation is necessary. Nevertheless, using the diagonal enables only parameter-wise estimates of sensitivity, as opposed to richer interactions between parameters in the full covariance matrix. Indeed, our results suggest that relying only on the diagonal is insufficient for maintaining safety, even with very high values of $\lambda$ to increase the strength of the regularisation. We argue in favour of adding an additional penalty: a Fisher projected onto the low-dimensional safety subspace, and therefore protecting it in a targeted way.

\subsection{\longmethod}
\label{sec:\longmethod}
Here, we first propose a method for protecting the subspace that spans safety-critical directions. We then consider different ways of operationalising the notion of a safety subspace, which give rise to different projection matrices, in Section \ref{sec:Locating the safety subspace}.

Our method builds on recent findings that safety-critical directions span a low-dimensional subspace. Recall that the diagonal FIM is computed by averaging element-wise squared gradients $g_j \in \mathbb{R}^{d_{\text{out}} \times d_{\text{in}}}$ of the loss with respect to a weight matrix $W_l$, evaluated at $\theta^{ref}$ on samples $j$ from $\mathcal{D}_{\text{Safety}}$. This avoids computing the full outer products $\mathrm{vec}(g_j)\mathrm{vec}(g_j)^\top$.

Assuming we have identified a suitable projection matrix $P_{r}$, we can project the gradients onto the $r$-dimensional safety subspace, yielding $\tilde{g}_j=P_r g_j \in \mathbb{R}^{r \times d_{\text{in}}}$. Averaging over the dataset gives the projected Fisher:

\begin{equation}
\label{eqn:projected FIM}
    \tilde{F} = \frac{1}{M}\sum_{j=1}^M \tilde{g}_j \tilde{g}_j^\top.
\end{equation}

Note that $\tilde F$ is not the full Fisher restricted to the subspace, which would still be of size $r d_\text{in} \times r d_\text{in}$ per layer and remain prohibitive at LLM scale. Instead, our approximation captures interactions among the $r$ output-side safety directions, aggregated over input dimensions.  This reduces the size of the Fisher approximation to $r \times r$ per layer. We also project the parameter changes $\Delta = \theta_l^t - \theta_l^{ref}$ onto the safety subspace, yielding $\tilde{\Delta}_l = P_r \Delta \in \mathbb{R}^{r \times d_{\text{in}}}$. We then penalise these changes using the projected Fisher:

\begin{equation}
\mathrm{tr}(\tilde{\Delta}_l^\top \tilde{F}_l \tilde{\Delta}_l),
\end{equation}

\noindent where $\tilde{\Delta}_l^\top \tilde{F}_l \tilde{\Delta}_l \in \mathbb{R}^{d_{\text{in}} \times d_{\text{in}}}$. Since $\tilde{F}$ is positive semi-definite, this penalty is always non-negative, and increases quadratically with parameter changes within the safety subspace. We further scale the penalty by $\lambda_\text{LR}$ as a controllable hyper-parameter. Summing across all linear layers 

and adding the penalty to the standard EWC objective yields the \method~objective:

\begin{equation}
    \mathcal{L}_\text{SP-EWC} = \mathcal{L}_\text{EWC} + \sum_i \frac{\lambda_\text{LR}}{2}\text{tr}( \tilde{\Delta}_i^\top \tilde{F}_i \tilde{\Delta}_i).
\end{equation}

Another perspective comes from considering the KL divergence between the reference model and the model during training,
$\text{KL}(p_{\theta^{ref}} || p_\theta)$ on safety data. Its second-order expansion is  $\frac{1}{2} \Delta^\top F \Delta$,  where $F$ is the Fisher information matrix. Both EWC and \method~aim to approximate this penalty to limit changes to the model predictions on safety data.   
While standard EWC uses the diagonal Fisher, \method~adds a penalty based on an approximation of the Fisher  restricted to the output-side safety directions. Parameter changes orthogonal to these directions are then penalised only by the standard EWC penalty term. This choice is motivated by findings that safety-relevant behaviour is concentrated in low-dimensional activation subspaces \citep{wei2024assessing}, while avoiding the cost of modelling interactions across both input and output dimensions.

\subsection{Locating the safety subspace}
\label{sec:Locating the safety subspace}

Having established a method for penalising updates along safety-critical directions, we now specify how to identify the safety subspace. We compare three approaches: 
low-rank adapters (Section \ref{sec:From low-rank adapters}), activation-based SVD (Section \ref{sec:From activations}), and contrastive PCA (Section \ref{sec:From contrastive principal component analysis}).

\subsubsection{Safety subspaces from LoRA adapters}
\label{sec:From low-rank adapters}
As a simple baseline, we use LoRA adapters \citep{hu2022lora} to identify the safety subspace. LoRA represents an update to a weight matrix $W_0 \in \mathbb{R}^{d_{\text{out}} \times d_{\text{in}}}$ as $\Delta W = BA$, where $B \in \mathbb{R}^{d_{\text{out}} \times r}, A \in \mathbb{R}^{r \times d_{\text{in}}}$. We train rank-$r$ adapters in every linear layer on safety data, starting from the reference model $\theta^{ref}$. For each layer, we then compute the SVD of $\Delta W = BA$ and use the top-$r$ left singular vectors $U^\top$ to construct the projection matrix $P_r = U^\top$. Intuitively, the directions associated with safety fine-tuning updates may identify a subspace relevant to preserving safety behaviour.

\subsubsection{Safety subspaces from activation SVD}
\label{sec:From activations}

Another  approach is to identify the safety subspace directions from model activations, following ActSVD \citep{wei2024assessing}. We collect input activations $X_{\text{in}}$ on $\mathcal{D}_{\text{Safety}}$ and compute the top-$r$ left singular vectors $U_s$ of the corresponding output activations $WX_{\text{in}}$. This yields the safety projection matrix $P_s = U_s U_s^\top$. However, these directions may also capture general language modelling capabilities rather than safety-specific behaviour. We may not want to protect these directions, as they need to be adjusted while training models for reasoning. Following \citet{wei2024assessing}, we therefore construct a utility projection matrix $P_u$ in the same way using $\mathcal{D}_{\text{Util}}$ and remove utility-related directions:

\begin{equation}
    \hat{W} = (I - P_u) P_s W.
\end{equation}
We then obtain the top-$r$ left singular vectors of $\hat W$ and use their transpose as the projection matrix $P_r$ in SP-EWC.

\subsubsection{Safety subspaces from contrastive PCA}
\label{sec:From contrastive principal component analysis}
Contrastive Principal Component Analysis (cPCA; \citealp{Abid2018-iq}), to our knowledge not previously applied as an intervention for protecting safety, offers an alternative approach.  Unlike ActSVD, which first identifies prominent directions in safety activations and then removes utility-related components, cPCA searches for directions with high variance in safety activations and low variance in utility activations. This has an important advantage: directions relevant to safety may also appear in utility activations and removing the utility subspace entirely could discard useful safety information, disabling its protection. By contrasting the two distributions, cPCA lets us control the trade-off between retaining safety-related variation and suppressing variation shared with utility data. 

As in ActSVD, we collect output activations $WX_{\text{in}}$ for each linear layer on both $\mathcal{D}_{\text{Safety}}$ and $\mathcal{D}_{\text{Util}}$. Rather than analysing the activation matrices separately, cPCA considers their covariance matrices $C_s$ and $C_u$,\footnote{Recall that PCA can equivalently be performed using SVD of centred data: the principal components are the left singular vectors of the centred activation matrix.} and identifies directions $\mathbf{v}$ that maximise
\begin{equation}
    \operatorname*{arg\,max}_{\|\mathbf{v}\|=1}
    \mathbf{v}^\top(C_s-\alpha C_u)\mathbf{v},
\end{equation}
where $\alpha \geq 0$ controls how strongly variation in utility activations is penalised. We use the top-$r$ eigenvectors of $C_s-\alpha C_u$ as the columns of $U$, yielding the projection matrix $P_r=U^\top$ for each layer. In our experiments, cPCA is the most effective of the three approaches, providing more consistent safety preservation across models.

\begin{figure*}[ht]
\begin{center}
\includegraphics[width=1\linewidth]{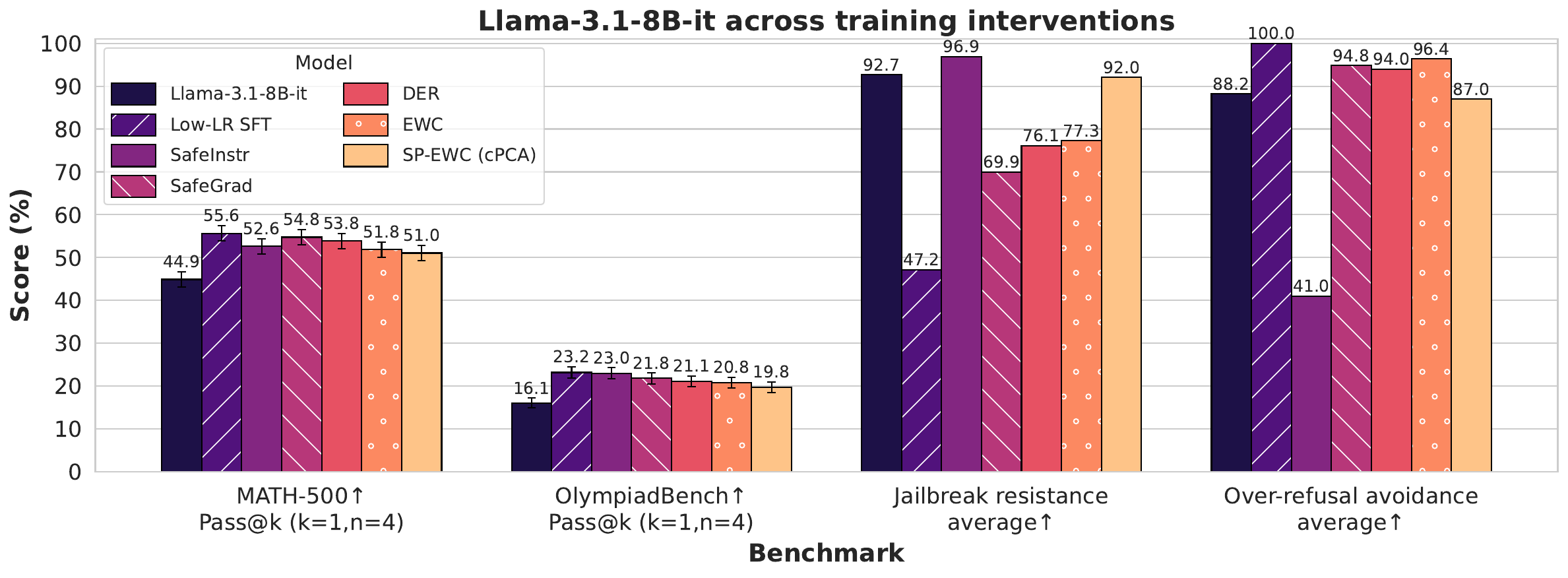}
\end{center}
\caption{The performance of Llama-3.1-8B-it and various interventions on MATH-500, OlympiadBench, its average on jailbreak resistance benchmarks, and its average on over-refusal benchmarks. Notably, SafeInstr over-optimises for jailbreak resistance, introducing over-refusal as a new pathology. All other baselines struggle to maintain safety guardrails, while \method~strikes a favourable trade-off. The full results table detailing individual safety benchmark results can be found in Table \ref{tab:Full results table}.}
\label{fig:Main results figure}
\end{figure*}

\section{Experimental setup}
\label{sec:Experimental setup}
\paragraph{Datasets} We train models on reasoning chains from subsets of the \texttt{open-r1/Mixture-of-Thoughts} dataset (details  in Appendix \ref{sec:Appendix - Further details on implementation}). We evaluate reasoning performance on MATH-500 \citep{hendrycks2021measuring, lightman2024lets} and OlympiadBench \citep{he-etal-2024-olympiadbench} with HuggingFace's \texttt{LightEval} benchmarking harness \citep{lighteval}.  For safety, we use AllenAI's \texttt{safety-eval} \citep{safety-eval}, with default settings, following \citet{beyer2026position}. We report two aggregate metrics: \textbf{jailbreak resistance (JR)}, averaged over DoAnythingNow \citep{inproceedings},  WildJailbreak (harmful) \citep{jiang2024wildteaming}, WildGuardTest (adversarial) \citep{han2024wildguard}, and TrustLLM (JailbreakTrigger) \citep{Huang2024TrustLLM}; and \textbf{over-refusal avoidance (OA)}, averaged over XSTest \citep{rottger-etal-2024-xstest} and WildJailbreak (benign) \citep{jiang2024wildteaming}. All metrics are reported with higher values indicating better performance: the proportion of safe responses for jailbreak resistance and correct compliance for over-refusal avoidance.

\paragraph{Models} We select models with sufficiently strong safety guardrails before fine-tuning, so that any degradation in safety can be meaningfully measured. We use two models: \texttt{Llama-3.1-8B-it} and \texttt{OLMo-2-7B-it}. We justify this choice via our preliminary results in Table \ref{tab:Choice of models}. Where the original safety training data is not available,  we construct our own safety dataset. Following \citet{xu2025magpie}, we sample  model completions for prompts from the safety split of WildGuardMix's training set \citep{han2024wildguard}, which is disjoint from our evaluation benchmarks. We keep completions judged to be safe by a safety judge \citep{han2024wildguard} and use them as $\mathcal{D}_{\text{Safety}}$ for all methods that require safety data. We use the same procedure to construct $\mathcal{D}_{\text{Util}}$, using only benign prompts from WildGuardMix. We provide further details in Appendix \ref{sec:Appendix - Further details on implementation}.

We compare against several methods. As the SFT baseline, we consider low learning rate fine-tuning \citep{kim2025rethinking, lin2026sft}, which has been found to help mitigate prior task degradation. Similarly, we compare against SafeInstr \citep{bianchi2024safetytuned}, for which we mix $\mathcal{D}_{\text{Safety}}$ into the training data. As a gradient surgery intervention, we consider SafeGrad \citep{yi2026safegrad}, which computes the KL divergence to the response distribution of $\theta^{ref}$ and performs gradient surgery in the case of gradient conflicts. To isolate the role of this surgery, we also consider a KL variant of DER \citep{Buzzega2020DER}, which \citet{alssum2025unforgottensafetypreservingsafety} found to compare favourably against other CL methods. Finally, to highlight the value of subspace projection, we compare against EWC \citep{elastic_weight_consolidation}.

\begin{figure*}[ht]
\begin{center}
\includegraphics[width=1\linewidth]{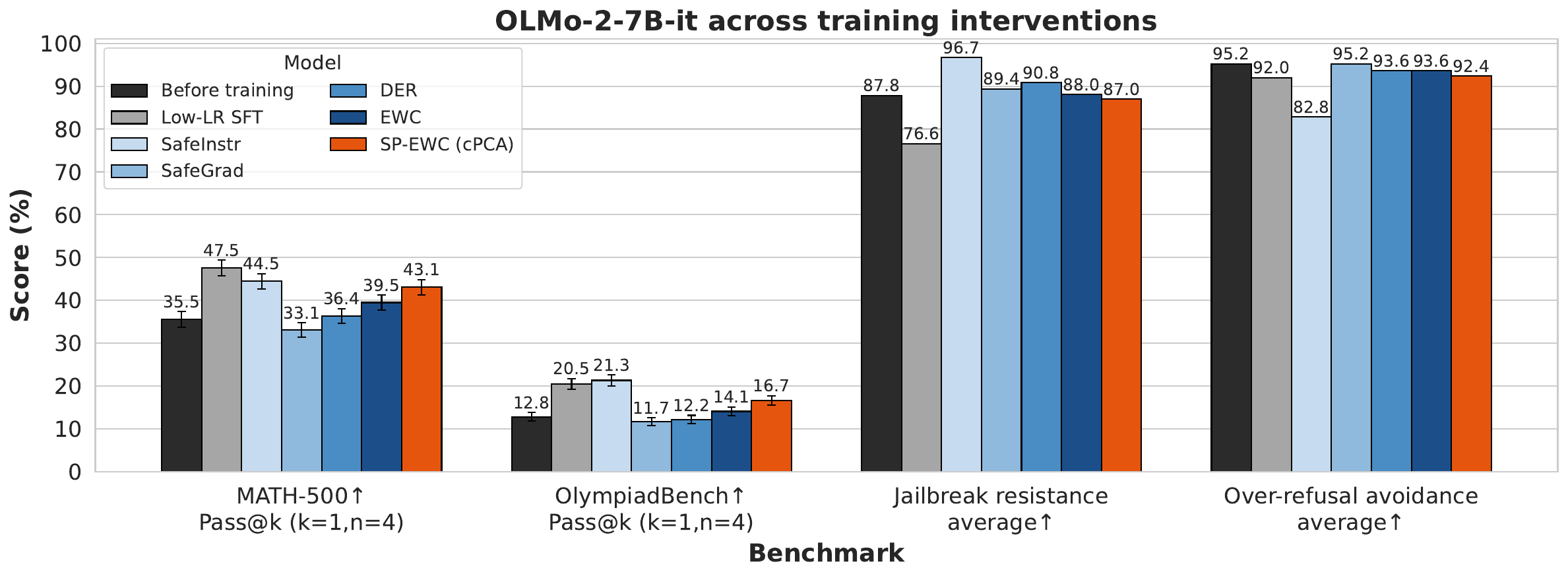}
\end{center}
\caption{The performance of OLMo-2-7B-it and various interventions on maths and safety benchmarks. Notably, SafeInstr over-optimises for jailbreak resistance, introducing over-refusal as a new pathology. \method~better balances this while still enabling in-domain gains. The full results table detailing individual safety benchmark results can be found in Table \ref{tab:Full results table}.}
\label{fig:Main results figure - olmo}
\end{figure*}

\section{Results}
\label{sec:Results}

\subsection{Main results}
\label{sec:Main results}

Figures \ref{fig:Main results figure} and \ref{fig:Main results figure - olmo} show our results for Llama-3.1-8B-it and OLMo-2-7B-it, respectively, with computational overhead in Table \ref{tab:compute overhead}.
Ablations are shown in appendices \ref{sec:Appendix - Varying the strength of the EWC penalty}, \ref{sec:Appendix - SafeGrad and DER under varying amounts of safety data}, \ref{sec:Appendix - Varying the learning rate}, and \ref{sec:Appendix - Varying the amount of safety data}.

Recent work has shown that reducing the learning rate during fine-tuning can substantially mitigate forgetting while retaining most of the gains on the new task \citep{kim2025rethinking, lin2026sft}. We investigate this in our setting (Table \ref{tab:Varying the learning rate}) and find that lowering the learning rate reduces safety degradation, but substantial losses in jailbreak resistance remain. We therefore use the lower learning rate in all subsequent experiments, ensuring that our methods are evaluated against a stronger baseline.

The primary objective of this setup is to mitigate degradations in JR across training. We observe SafeInstr to do this best, appearing even safer than Llama-3.1-8B-it before training. However, it also introduces a massive amount of over-refusal. This replicates findings by \citet{bianchi2024safetytuned}, and shows the necessity of balancing strong guardrails with continued utility on benign prompts. We ablate safety data percentage in Table \ref{tab:Varying the amount of safety data}.

\begin{figure*}[!b]
\begin{center}
\includegraphics[width=1\linewidth]{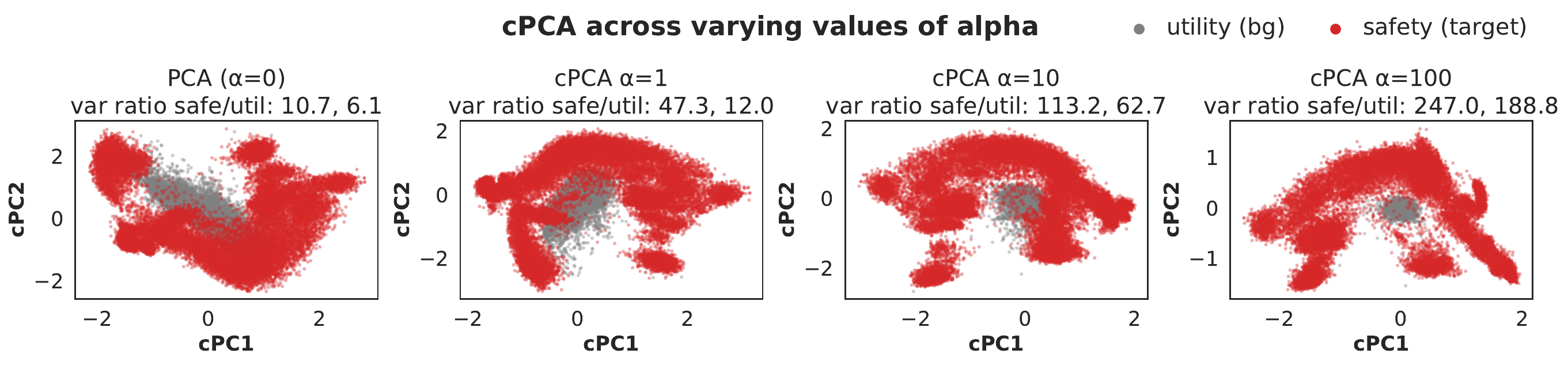}
\end{center}
\caption{We plot cPCA for a single linear layer to illustrate its effects. We collect activations from the MLP down projection of layer 16 of Llama-3.1-8B-it, and project them onto the top two contrastive principal components. Activations come from the first 20 response tokens of 2000 safety (target) and 2000 utility (background) examples. As $\alpha$ rises, the principal components exhibit comparatively more variance in safety than utility activations.}
\label{fig:cPCA figure}
\end{figure*}

Furthermore, we find that while SafeGrad and DER both capture much of the maths reasoning gain of standard SFT in Llama-3.1-8B-it, they fail to fully maintain the JR of the base model. This works better when used on OLMo-2-7B-it, though comes at a greater cost in MATH-500 and OlympiadBench performance: When using SafeGrad, the model is worse at both, and DER seems to stagnate. EWC provides some uplift, with improvements on JR in Llama-3.1-8B-it, but still fails to balance all three objectives effectively.

In contrast, we find that \method~with cPCA maintains JR in Llama-3.1-8B-it well, but without introducing over-refusal as a new pathology like SafeInstr. When training OLMo-2-7B-it, it recovers maths reasoning gains better than other regularisation-based methods, while maintaining JR effectively and avoiding over-refusals. It still comes at a cost: Llama-3.1-8B-it sees a reduction in MATH-500 and OlympiadBench improvements in contrast to tuned baselines. However, SP-EWC achieves a good balance between preserving jailbreak resistance, avoiding over-refusal, and improving reasoning performance.

\begin{figure}[h]
\begin{center}
\includegraphics[width=1\linewidth]{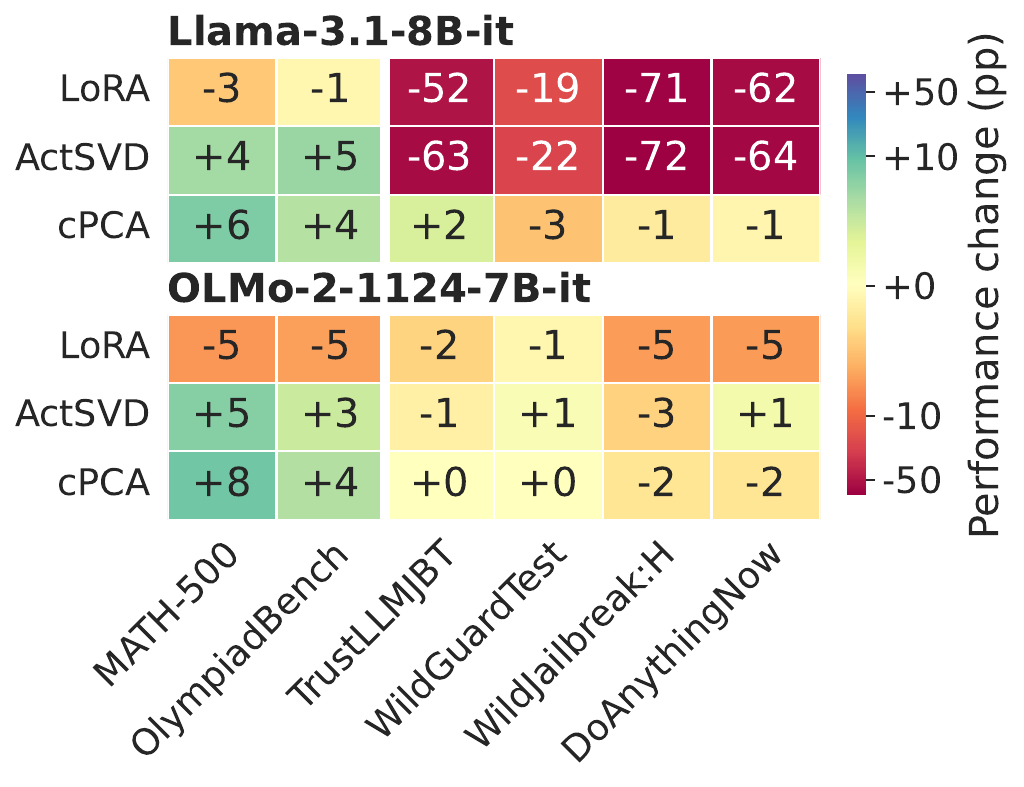}
\end{center}
\caption{Comparison of the performance change in percentage points when locating the safety subspace for \method~via LoRA, ActSVD or cPCA.}
\label{fig:Subspace comparison figure}
\end{figure}

\subsection{Ablating the role of the projected Fisher}

\begin{table}[h]
    \centering
    \small
    \begin{tabular}{l *{4}{c}}
        \toprule
        Variant & M-500 & Olymp & JR & OA \\
        \midrule
        \textit{Llama-3.1-8B-it} & \textit{44.9} & \textit{16.1} & \textit{92.7} & \textit{88.2} \\
        Diagonal only & 51.8 & 20.8 & 77.3 & 96.4 \\
        Diagonal + subspace & 51.0 & 19.8 & 92.0 & 87.0 \\
        Subspace only & 50.6 & 21.8 & 89.9 & 87.2 \\
        \midrule
        \textit{OLMo-2-7B-it} & \textit{35.5} & \textit{12.8} & \textit{87.8} & \textit{95.2} \\
        Diagonal only & 39.5 & 14.1 & 88.0 & 93.6 \\
        Diagonal + subspace & 43.1 & 16.7 & 87.0 & 92.4 \\
        Subspace only & 46.4 & 18.1 & 84.7 & 93.0 \\
        \bottomrule
    \end{tabular}
    \caption{Ablation separating the contributions of the diagonal EWC penalty and the projected Fisher penalty in \method~(cPCA).}
    \label{tab:Ablating the diagonal and subspace penalties}
\end{table}

To investigate the role of the diagonal and subspace penalties, we provide an ablation in Table \ref{tab:Ablating the diagonal and subspace penalties}. We find that combining the diagonal with the subspace maintains jailbreak resistance well, while still enabling in-domain gains. Trends are not uniform across models: for Llama it is critical for maintaining JR, and for OLMo it helps recover more of the in-domain gain.

\subsection{Comparing methods for subspaces}
\label{sec:Comparing methods for subspaces}

In order to understand the role of the subspace location method, we ablate this choice and compare the impact of using \method~with LoRA, ActSVD and cPCA. Our results are shown in Figure \ref{fig:Subspace comparison figure}.

We find that LoRA seems to not only hinder the learning of the core task, with slight degradations on MATH-500 and OlympiadBench, but also degrades the performance of the model across all four JR benchmarks we tested. 
One possible explanation is that safety training is already close to saturation in the reference model, so further fine-tuning with LoRA may capture directions associated with adapting to the new safety data rather than those responsible for the model's existing safety guardrails.

Meanwhile, ActSVD and cPCA yield better results for OLMo-2-7B-it, with virtually the same gains on MATH-500 and OlympiadBench, and very similar preservation of JR. However, \method~performs poorly on Llama-3.1-8B-it when using ActSVD, causing drops in JR. We speculate that OLMo-2-7B-it may have more robust safety guardrails, as suggested by the smaller degradation in JR under low-learning-rate SFT. Furthermore, ActSVD introduces two hyper-parameters, the safety and utility ranks, which may interact in complex ways: projecting out too many utility-related directions could leave a subspace that is uninformative for safety. In contrast, cPCA appears to be more robust across models.

To better understand why cPCA identifies the safety subspace effectively, we project activations from 2,000 safety and 2,000 utility prompts onto the top two contrastive principal components while varying $\alpha$. The results are shown in Figure \ref{fig:cPCA figure}. When $\alpha=0$, cPCA reduces to standard PCA, and the projections of the two datasets overlap substantially. As $\alpha$ increases, cPCA favours directions with high variance in safety activations and low variance in utility activations \citep{Abid2018-iq}. At $\alpha=100$, the variance along the two components is much higher for safety than for utility data. While such two-dimensional projections should be interpreted cautiously, these observations are consistent with cPCA identifying more safety-specific directions, potentially explaining its stronger downstream performance compared to ActSVD.

\section{Discussion}
\label{sec:Discussion}

\paragraph{Locating the safety subspace correctly is key for protecting guardrails}
Prior work has established that safety guardrails exist within a low-rank subspace within models, and perturbations to it can remove them \citep{peng2024navigating, wei2024assessing}. Our observations suggest that the identification of this subspace also enables protecting it in a targeted way. Results in Section \ref{sec:Comparing methods for subspaces} indicate that choosing the right method for locating the subspace is critical: Doing so via LoRA adapters actively hindered task learning, while severely damaging the safety guardrails trained into the model by its developers.

\paragraph{When capabilities exist in low-dimensional subspaces, SP-EWC may enable their protection}
SP-EWC can be extended beyond the safety domain. For other capabilities associated with low-dimensional subspaces, approximating the Fisher within these subspaces could enable their protection during fine-tuning via EWC. This is a promising direction that we leave for future work.

\label{sec:Locating the safety subspace correctly is key for protecting guardrails}
\paragraph{What does it mean to maintain safety during fine-tuning?}
Our approach builds on findings that safety-related behaviour can be controlled through low-dimensional subspaces. However, this focuses on refusals, rather than safety in a broader sense. Importantly, classifying outputs as ``safe'' or ``unsafe'' depends on a policy \citep{openaiOpenAIModel, anthropicClaudesConstitution} that is typically written by model developers, as opposed to democratic institutions \citep{hernandez-cano-etal-2026-apertus}. Furthermore, this view excludes concerns such as environmental impacts \citep{Luccioni2024Power, Luccioni2025Jevons, jegham2025hungryaibenchmarkingenergy, morrison2025holistically}, intellectual property theft \citep{mammen2024a, collett_2025, JMLR:v24:23-0569}, or cybersecurity concerns \citep{openaiHuggingFace, metr-2026-openai-hugging-face-incident-investigation}, among others. Finally, safety is a probabilistic problem: a model with a low ASR may still generate unsafe responses under repeated sampling \citep{angell2026estimating}. We therefore emphasise that ensuring model safety extends beyond preserving existing guardrails as measured by benchmarks. Our results should be interpreted in the broader context of LLM development and deployment.

\section{Conclusion}
\label{sec:Conclusion}
In this paper, we investigate the problem of benign fine-tuning deteriorating safety guardrails in LLMs. We exploit recent findings that safety is low-rank by modifying EWC to project the Fisher onto a safety subspace identified with cPCA. We find that this protects the safety capabilities of LLMs, balancing jailbreak resistance and over-refusal while still teaching models to reason.

\section*{Limitations}
\label{sec:Limitations}
Our work operationalises safety guardrails via jailbreak and refusal benchmarks that are widely used in the field. However, its scope does not extend to other important areas of safety like bias, multilingual behaviour or long-horizon tasks. Furthermore, we focus on reasoning training via SFT, which is an important setting as SFT is a central way for teaching models new capabilities. Investigating impacts during RL is an interesting direction that we leave for future work. Finally, while the computational overhead compared to standard EWC is negligible, \method~still introduces a higher cost than cheaper methods like SafeInstr.

\section*{Acknowledgments}
Neel Rajani is funded by the UKRI AI Centre for Doctoral Training in Responsible and Trustworthy in-the-world Natural Language Processing (grant ref: EP/Y030656/1). Ivan Titov acknowledges the support of the Netherlands Organization for Scientific Research, NWO (VICI grant VI.C.212.053). Computational resources were provided by the Edinburgh International Data Facility (EIDF) and the Data-Driven Innovation Programme at the University of Edinburgh. Access to EIDF was facilitated through the University of Edinburgh’s Generative AI Laboratory GAIL Fellow scheme. We thank Cesare Spinoso-Di Piano, Pasquale Minervini and Aryo Pradipta Gema for feedback on earlier drafts.

In this work, we used generative AI tools (mainly Claude Opus 5.5 in Claude Code) for assistance in surfacing papers following our own literature review, help in debugging, and for some boilerplate code. We wrote the paper ourselves, with limited LLM assistance for language polishing, catching errors and feedback. We have not used generative AI tools for tasks beyond those disclosed above.

\bibliography{custom}

@inproceedings{
    qi2024finetuning,
    title={Fine-tuning Aligned Language Models Compromises Safety, Even When Users Do Not Intend To!},
    author={Xiangyu Qi and Yi Zeng and Tinghao Xie and Pin-Yu Chen and Ruoxi Jia and Prateek Mittal and Peter Henderson},
    booktitle={The Twelfth International Conference on Learning Representations},
    year={2024},
    url={https://openreview.net/forum?id=hTEGyKf0dZ}
    }

@misc{semianalysisLongLive,
	author = {Myron Xie},
	title = {{L}ong {L}ive the {S}hort {K}ing: {W}hy 4-hi {H}{B}{M} {W}ins --- newsletter.semianalysis.com},
	howpublished = {\url{https://newsletter.semianalysis.com/p/long-live-the-short-king-why-4-hi}},
	year = {2026},
	note = {[Accessed 16-09-2026]},
    }

@misc{openaiHuggingFace,
	author = {OpenAI},
	title = {{T}he {H}ugging {F}ace incident and the road ahead --- openai.com},
	howpublished = {\url{https://openai.com/index/hugging-face-incident-and-the-road-ahead/}},
	year = {2026},
	note = {[Accessed 22-09-2026]},
}

@misc{metr-2026-openai-hugging-face-incident-investigation,
    title = {Brief independent investigation of agents’ behavior, reasoning and collaboration in the OpenAI / Hugging Face hacking incident},
    author = {METR},
    howpublished = {\url{https://metr.org/blog/2026-08-26-openai-hugging-face-incident-investigation/}},
    year = {2026},
    month = {08},
}

@inproceedings{
    arditi2024refusal,
    title={Refusal in Language Models Is Mediated by a Single Direction},
    author={Andy Arditi and Oscar Balcells Obeso and Aaquib Syed and Daniel Paleka and Nina Rimsky and Wes Gurnee and Neel Nanda},
    booktitle={The Thirty-eighth Annual Conference on Neural Information Processing Systems},
    year={2024},
    url={https://openreview.net/forum?id=pH3XAQME6c}
    }

@inproceedings{
    peng2024navigating,
    title={Navigating the Safety Landscape: Measuring Risks in Finetuning Large Language Models},
    author={ShengYun Peng and Pin-Yu Chen and Matthew Daniel Hull and Duen Horng Chau},
    booktitle={The Thirty-eighth Annual Conference on Neural Information Processing Systems},
    year={2024},
    url={https://openreview.net/forum?id=GZnsqBwHAG}
    }

@inproceedings{
    hu2022lora,
    title={Lo{RA}: Low-Rank Adaptation of Large Language Models},
    author={Edward J Hu and yelong shen and Phillip Wallis and Zeyuan Allen-Zhu and Yuanzhi Li and Shean Wang and Lu Wang and Weizhu Chen},
    booktitle={International Conference on Learning Representations},
    year={2022},
    url={https://openreview.net/forum?id=nZeVKeeFYf9}
    }

@ARTICLE{Abid2018-iq,
  title     = "Exploring patterns enriched in a dataset with contrastive
               principal component analysis",
  author    = "Abid, Abubakar and Zhang, Martin J and Bagaria, Vivek K and Zou,
               James",
  journal   = "Nat. Commun.",
  publisher = "Springer Science and Business Media LLC",
  volume    =  9,
  number    =  1,
  pages     = "2134",
  month     =  may,
  year      =  2018,
  copyright = "https://creativecommons.org/licenses/by/4.0",
  language  = "en"
}

@misc{jiang2025safechainsafetylanguagemodels,
      title={SafeChain: Safety of Language Models with Long Chain-of-Thought Reasoning Capabilities}, 
      author={Fengqing Jiang and Zhangchen Xu and Yuetai Li and Luyao Niu and Zhen Xiang and Bo Li and Bill Yuchen Lin and Radha Poovendran},
      year={2025},
      eprint={2502.12025},
      archivePrefix={arXiv},
      primaryClass={cs.AI},
      url={https://arxiv.org/abs/2502.12025}, 
}

@misc{yong2026selfjailbreakinglanguagemodelsreason,
      title={Self-Jailbreaking: Language Models Can Reason Themselves Out of Safety Alignment After Benign Reasoning Training}, 
      author={Zheng-Xin Yong and Stephen H. Bach},
      year={2026},
      eprint={2510.20956},
      archivePrefix={arXiv},
      primaryClass={cs.CR},
      url={https://arxiv.org/abs/2510.20956}, 
}

@misc{zhou2025hiddenriskslargereasoning,
      title={The Hidden Risks of Large Reasoning Models: A Safety Assessment of R1}, 
      author={Kaiwen Zhou and Chengzhi Liu and Xuandong Zhao and Shreedhar Jangam and Jayanth Srinivasa and Gaowen Liu and Dawn Song and Xin Eric Wang},
      year={2025},
      eprint={2502.12659},
      archivePrefix={arXiv},
      primaryClass={cs.CY},
      url={https://arxiv.org/abs/2502.12659}, 
}

@inproceedings{
    li2025finding,
    title={Finding and Reactivating Post-Trained {LLM}s' Hidden Safety Mechanisms},
    author={Mingjie Li and Wai Man Si and Michael Backes and Yang Zhang and Yisen Wang},
    booktitle={The Thirty-ninth Annual Conference on Neural Information Processing Systems},
    year={2025},
    url={https://openreview.net/forum?id=pcG6NRJKu7}
    }

@inproceedings{
    beyer2026position,
    title={Position: {LLM}-Safety Evaluations Lack Robustness},
    author={Tim Beyer and Sophie Xhonneux and Simon Geisler and Gauthier Gidel and Leo Schwinn and Stephan G{\"u}nnemann},
    booktitle={Forty-third International Conference on Machine Learning Position Paper Track},
    year={2026},
    url={https://openreview.net/forum?id=npJ7tn2QO8}
    }

@inproceedings{huang2024Lisa,
    author = {Huang, Tiansheng and Hu, Sihao and Ilhan, Fatih and Tekin, Selim Furkan and Liu, Ling},
    title = {Lisa: lazy safety alignment for large language models against harmful fine-tuning attack},
    year = {2024},
    isbn = {9798331314385},
    publisher = {Curran Associates Inc.},
    address = {Red Hook, NY, USA},
    booktitle = {Proceedings of the 38th International Conference on Neural Information Processing Systems},
    articleno = {3320},
    numpages = {35},
    location = {Vancouver, BC, Canada},
    series = {NeurIPS '24}
    }

@inproceedings{
    lin2026sft,
    title={{SFT} Doesn{\textquoteright}t Always Hurt General Capabilities: Revisiting Domain-Specific Fine-Tuning in {LLM}s},
    author={Jiacheng Lin and Zhongruo Wang and Kun Qian and Tian Wang and Arvind Srinivasan and Hansi Zeng and Ruochen Jiao and Xie Zhou and Jiri Gesi and Dakuo Wang and Yufan Guo and Kai Zhong and Weiqi Zhang and sujay sanghavi and Changyou Chen and Hyokun Yun and Lihong Li},
    booktitle={The Fourteenth International Conference on Learning Representations},
    year={2026},
    url={https://openreview.net/forum?id=ru93xpQFi1}
    }

@article{
    elastic_weight_consolidation,
    author = {James Kirkpatrick  and Razvan Pascanu  and Neil Rabinowitz  and Joel Veness  and Guillaume Desjardins  and Andrei A. Rusu  and Kieran Milan  and John Quan  and Tiago Ramalho  and Agnieszka Grabska-Barwinska  and Demis Hassabis  and Claudia Clopath  and Dharshan Kumaran  and Raia Hadsell },
    title = {Overcoming catastrophic forgetting in neural networks},
    journal = {Proceedings of the National Academy of Sciences},
    volume = {114},
    number = {13},
    pages = {3521-3526},
    year = {2017},
    doi = {10.1073/pnas.1611835114},
    URL = {https://www.pnas.org/doi/abs/10.1073/pnas.1611835114},
    eprint = {https://www.pnas.org/doi/pdf/10.1073/pnas.1611835114}
    }

@inproceedings{
    li2025fishers,
    title={Fishers for Free? Approximating the Fisher Information Matrix by Recycling the Squared Gradient Accumulator},
    author={Yu Xin Li and Felix Dangel and Derek Tam and Colin Raffel},
    booktitle={Forty-second International Conference on Machine Learning},
    year={2025},
    url={https://openreview.net/forum?id=m3zrHhiCCj}
    }

@article{
    Note2018Huszar,
    author = {Ferenc Huszár },
    title = {Note on the quadratic penalties in elastic weight consolidation},
    journal = {Proceedings of the National Academy of Sciences},
    volume = {115},
    number = {11},
    pages = {E2496-E2497},
    year = {2018},
    doi = {10.1073/pnas.1717042115},
    URL = {https://www.pnas.org/doi/abs/10.1073/pnas.1717042115},
    eprint = {https://www.pnas.org/doi/pdf/10.1073/pnas.1717042115}
    }

@article{
    Reply2018Kirkpatrick,
    author = {James Kirkpatrick  and Razvan Pascanu  and Neil Rabinowitz  and Joel Veness  and Guillaume Desjardins  and Andrei A. Rusu  and Kieran Milan  and John Quan  and Tiago Ramalho  and Agnieszka Grabska-Barwinska  and Demis Hassabis  and Claudia Clopath  and Dharshan Kumaran  and Raia Hadsell },
    title = {Reply to Huszár: The elastic weight consolidation penalty is empirically valid},
    journal = {Proceedings of the National Academy of Sciences},
    volume = {115},
    number = {11},
    pages = {E2498-E2498},
    year = {2018},
    doi = {10.1073/pnas.1800157115},
    URL = {https://www.pnas.org/doi/abs/10.1073/pnas.1800157115},
    eprint = {https://www.pnas.org/doi/pdf/10.1073/pnas.1800157115}
    }

@inproceedings{limitations2019,
    title = {Limitations of the empirical fisher approximation for natural gradient descent},
    year = {2019},
    author={Frederik Kunstner and Lukas Balles and Philipp Hennig},
    publisher = {Curran Associates Inc.},
    address = {Red Hook, NY, USA},
    booktitle = {Proceedings of the 33rd International Conference on Neural Information Processing Systems},
    articleno = {374},
    numpages = {12}
    }

@inproceedings{
    peirson2022fishy,
    title={Fishy: Layerwise Fisher Approximation for Higher-order Neural Network Optimization},
    author={Abel Peirson and Ehsan Amid and Yatong Chen and Vladimir Feinberg and Manfred K Warmuth and Rohan Anil},
    booktitle={Has it Trained Yet? NeurIPS 2022 Workshop},
    year={2022},
    url={https://openreview.net/forum?id=cScb-RrBQC}
    }

@inproceedings{
    martens2018kroneckerfactored,
    title={Kronecker-factored Curvature Approximations for Recurrent Neural Networks},
    author={James Martens and Jimmy Ba and Matt Johnson},
    booktitle={International Conference on Learning Representations},
    year={2018},
    url={https://openreview.net/forum?id=HyMTkQZAb},
    }

@article{Schioppa2022,
  title = {Scaling Up Influence Functions},
  volume = {36},
  ISSN = {2159-5399},
  url = {http://dx.doi.org/10.1609/aaai.v36i8.20791},
  DOI = {10.1609/aaai.v36i8.20791},
  number = {8},
  journal = {Proceedings of the AAAI Conference on Artificial Intelligence},
  publisher = {Association for the Advancement of Artificial Intelligence (AAAI)},
  author = {Schioppa,  Andrea and Zablotskaia,  Polina and Vilar,  David and Sokolov,  Artem},
  year = {2022},
  month = {June},
  pages = {8179–8186}
}

@InProceedings{pmlr-v80-gupta18a,
  title = 	 {Shampoo: Preconditioned Stochastic Tensor Optimization},
  author =       {Gupta, Vineet and Koren, Tomer and Singer, Yoram},
  booktitle = 	 {Proceedings of the 35th International Conference on Machine Learning},
  pages = 	 {1842--1850},
  year = 	 {2018},
  editor = 	 {Dy, Jennifer and Krause, Andreas},
  volume = 	 {80},
  series = 	 {Proceedings of Machine Learning Research},
  month = 	 {10--15 Jul},
  publisher =    {PMLR},
  url = 	 {https://proceedings.mlr.press/v80/gupta18a.html}
}

@inproceedings{NEURIPS2018_f31b2046,
 author = {Ritter, Hippolyt and Botev, Aleksandar and Barber, David},
 booktitle = {Advances in Neural Information Processing Systems},
 editor = {S. Bengio and H. Wallach and H. Larochelle and K. Grauman and N. Cesa-Bianchi and R. Garnett},
 pages = {},
 publisher = {Curran Associates, Inc.},
 title = {Online Structured Laplace Approximations for Overcoming Catastrophic Forgetting},
 url = {https://proceedings.neurips.cc/paper_files/paper/2018/file/f31b20466ae89669f9741e047487eb37-Paper.pdf},
 volume = {31},
 year = {2018}
}

@inproceedings{
    ven2025on,
    title={On the Computation of the Fisher Information in Continual Learning},
    author={Gido M van de Ven},
    booktitle={The Fourth Blogpost Track at ICLR 2025},
    year={2025},
    url={https://openreview.net/forum?id=mFzh95dClk}
    }

@inproceedings{
    zheng2026revisiting,
    title={Revisiting Weight Regularization for Low-Rank Continual Learning},
    author={Yaoyue Zheng and Yin Zhang and Joost van de Weijer and Gido M van de Ven and Shaoyi Du and Xuetao Zhang and Zhiqiang Tian},
    booktitle={The Fourteenth International Conference on Learning Representations},
    year={2026},
    url={https://openreview.net/forum?id=pZj2DhfaVD}
    }

@inproceedings{Xiang2023EWC-world-models,
    author = {Xiang, Jiannan and Tao, Tianhua and Gu, Yi and Shu, Tianmin and Wang, Zirui and Yang, Zichao and Hu, Zhiting},
    title = {Language models meet world models: embodied experiences enhance language models},
    year = {2023},
    publisher = {Curran Associates Inc.},
    address = {Red Hook, NY, USA},
    booktitle = {Proceedings of the 37th International Conference on Neural Information Processing Systems},
    articleno = {3295},
    numpages = {21},
    location = {New Orleans, LA, USA},
    series = {NIPS '23}
    }

@misc{sliogeris2025fullparametercontinualpretraininggemma2,
      title={Full-Parameter Continual Pretraining of Gemma2: Insights into Fluency and Domain Knowledge}, 
      author={Vytenis Šliogeris and Povilas Daniušis and Artūras Nakvosas},
      year={2025},
      eprint={2505.05946},
      archivePrefix={arXiv},
      primaryClass={cs.CL},
      url={https://arxiv.org/abs/2505.05946}, 
    }

@inproceedings{Buzzega2020DER,
     author = {Buzzega, Pietro and Boschini, Matteo and Porrello, Angelo and Abati, Davide and CALDERARA, SIMONE},
     booktitle = {Advances in Neural Information Processing Systems},
     editor = {H. Larochelle and M. Ranzato and R. Hadsell and M.F. Balcan and H. Lin},
     pages = {15920--15930},
     publisher = {Curran Associates, Inc.},
     title = {Dark Experience for General Continual Learning: a Strong, Simple Baseline},
     url = {https://proceedings.neurips.cc/paper_files/paper/2020/file/b704ea2c39778f07c617f6b7ce480e9e-Paper.pdf},
     volume = {33},
     year = {2020}
    }

@inproceedings{wei2025online,
  title={Online-lora: Task-free online continual learning via low rank adaptation},
  author={Wei, Xiwen and Li, Guihong and Marculescu, Radu},
  booktitle={2025 IEEE/CVF Winter Conference on Applications of Computer Vision (WACV)},
  pages={6634--6645},
  year={2025},
  organization={IEEE}
}

@inproceedings{
    qi2025safety,
    title={Safety Alignment Should be Made More Than Just a Few Tokens Deep},
    author={Xiangyu Qi and Ashwinee Panda and Kaifeng Lyu and Xiao Ma and Subhrajit Roy and Ahmad Beirami and Prateek Mittal and Peter Henderson},
    booktitle={The Thirteenth International Conference on Learning Representations},
    year={2025},
    url={https://openreview.net/forum?id=6Mxhg9PtDE}
    }

@inproceedings{
    hsu2024safe,
    title={Safe Lo{RA}: The Silver Lining of Reducing Safety Risks when Finetuning Large Language Models},
    author={Chia-Yi Hsu and Yu-Lin Tsai and Chih-Hsun Lin and Pin-Yu Chen and Chia-Mu Yu and Chun-Ying Huang},
    booktitle={The Thirty-eighth Annual Conference on Neural Information Processing Systems},
    year={2024},
    url={https://openreview.net/forum?id=HcifdQZFZV}
    }

@inproceedings{
    wei2024assessing,
    title={Assessing the Brittleness of Safety Alignment via Pruning and Low-Rank Modifications},
    author={Boyi Wei and Kaixuan Huang and Yangsibo Huang and Tinghao Xie and Xiangyu Qi and Mengzhou Xia and Prateek Mittal and Mengdi Wang and Peter Henderson},
    booktitle={ICLR 2024 Workshop on Mathematical and Empirical Understanding of Foundation Models},
    year={2024},
    url={https://openreview.net/forum?id=niBPvgJIHB}
    }

@inproceedings{
    perin2025lox,
    title={LoX: Low-Rank Extrapolation Robustifies {LLM} Safety Against Fine-tuning},
    author={Gabriel Jacob Perin and Runjin Chen and Xuxi Chen and Nina S. T. Hirata and Zhangyang Wang and Junyuan Hong},
    booktitle={Second Conference on Language Modeling},
    year={2025},
    url={https://openreview.net/forum?id=ASS5YD4hL4}
    }

@inproceedings{
    yang2025asft,
    title={As{FT}: Anchoring Safety During {LLM} Fine-Tuning Within Narrow Safety Basin},
    author={Shuo Yang and Qihui Zhang and Yuyang Liu and Yue Huang and Xiaojun Jia and Kun-Peng Ning and Jia-Yu Yao and jigang wang and Dai Hailiang and Yibing Song and Li Yuan},
    booktitle={NeurIPS 2025 Workshop: Reliable ML from Unreliable Data},
    year={2025},
    url={https://openreview.net/forum?id=96UEsasTfE}
    }

@inproceedings{
    bianchi2024safetytuned,
    title={Safety-Tuned {LL}a{MA}s: Lessons From Improving the Safety of Large Language Models that Follow Instructions},
    author={Federico Bianchi and Mirac Suzgun and Giuseppe Attanasio and Paul Rottger and Dan Jurafsky and Tatsunori Hashimoto and James Zou},
    booktitle={The Twelfth International Conference on Learning Representations},
    year={2024},
    url={https://openreview.net/forum?id=gT5hALch9z}
    }

@INPROCEEDINGS{yi2026safegrad,
  author={Yi, Biao and Li, Jiahao and Zhang, Baolei and Nie, Lihai and Li, Tong and Huang, Tiansheng and Liu, Zheli},
  booktitle={ICASSP 2026 - 2026 IEEE International Conference on Acoustics, Speech and Signal Processing (ICASSP)}, 
  title={SafeGrad: Gradient Surgery for Safe LLM Fine-Tuning}, 
  year={2026},
  volume={},
  number={},
  pages={3326-3330},
  doi={10.1109/ICASSP55912.2026.11463951}
  }

@inproceedings{peng-etal-2026-preventing,
    title = "Preventing Safety Drift in Large Language Models via Coupled Weight and Activation Constraints",
    author = "Peng, Songping  and
      Zhang, Zhiheng  and
      Zeng, Daojian  and
      Jiang, Lincheng  and
      Gao, Xieping",
    editor = "Liakata, Maria  and
      Moreira, Viviane P.  and
      Zhang, Jiajun  and
      Jurgens, David",
    booktitle = "Findings of the {A}ssociation for {C}omputational {L}inguistics: {ACL} 2026",
    month = jul,
    year = "2026",
    address = "San Diego, California, United States",
    publisher = "Association for Computational Linguistics",
    url = "https://aclanthology.org/2026.findings-acl.874/",
    doi = "10.18653/v1/2026.findings-acl.874",
    pages = "17646--17662",
    ISBN = "979-8-89176-395-1"
}

@misc{alssum2025unforgottensafetypreservingsafety,
      title={Unforgotten Safety: Preserving Safety Alignment of Large Language Models with Continual Learning}, 
      author={Lama Alssum and Hani Itani and Hasan Abed Al Kader Hammoud and Philip Torr and Adel Bibi and Bernard Ghanem},
      year={2025},
      eprint={2512.10150},
      archivePrefix={arXiv},
      primaryClass={cs.CL},
      url={https://arxiv.org/abs/2512.10150}, 
}

@misc{fang2026grsapgenerativereplaysafety,
      title={GR-SAP: Generative Replay for Safety Alignment Preservation during Fine-Tuning}, 
      author={Zhouxiang Fang and Jiawei Zhou and Hanjie Chen},
      year={2026},
      eprint={2603.10243},
      archivePrefix={arXiv},
      primaryClass={cs.CL},
      url={https://arxiv.org/abs/2603.10243}, 
}

@inproceedings{
    angell2026estimating,
    title={Estimating Tail Risks in Language Model Output Distributions},
    author={Rico Angell and Raghav Singhal and Zachary Horvitz and Zhou Yu and Rajesh Ranganath and Kathleen McKeown and He He},
    booktitle={Forty-third International Conference on Machine Learning},
    year={2026},
    url={https://openreview.net/forum?id=Joka19sTny}
    }

@misc{anthropicClaudesConstitution,
	author = {Amanda Askell and Joe Carlsmith and Chris Olah and Jared Kaplan and Holden Karnofsky and Kyle Fish and Jack Lindsey and Nick Sofroniew and Evan Hubinger},
	title = {{C}laude’s {C}onstitution --- anthropic.com},
	howpublished = {\url{https://www.anthropic.com/constitution}},
	year = {2026},
	note = {[Accessed 14-09-2026]},
}

@misc{openaiOpenAIModel,
	author = {OpenAI},
	title = {{O}pen{A}{I} {M}odel {S}pec --- model-spec.openai.com},
	howpublished = {\url{https://model-spec.openai.com/2026-08-18.html}},
	year = {2026},
	note = {[Accessed 14-09-2026]},
}

@inproceedings{hernandez-cano-etal-2026-apertus,
    title = "Apertus: Democratizing Open and Compliant {LLM}s for Global Language Environments",
    author = {Hern{\'a}ndez-Cano, Alejandro  and
      H{\"a}gele, Alexander  and
      Huang, Allen Hao  and
      Romanou, Angelika  and
      Solergibert, Antoni-Joan  and
      P{\'a}sztor, Barna  and
      Messmer, Bettina  and
      Garbaya, Dhia  and
      {\v{D}}urech, Eduard Frank  and
      Hakimi, Ido  and
      Giraldo, Juan Garcia  and
      Ismayilzada, Mete  and
      Foroutan, Negar  and
      Moalla, Skander  and
      Chen, Tiancheng  and
      Sabol{\v{c}}ec, Vinko  and
      Xu, Yixuan  and
      Aerni, Michael  and
      AlKhamissi, Badr  and
      Marinas, In{\'e}s Altemir  and
      Amani, Mohammad Hossein  and
      Ansaripour, Matin  and
      Badanin, Ilia  and
      Benoit, Harold  and
      Boros, Emanuela  and
      Browning, Nicholas John  and
      B{\"o}sch, Fabian  and
      B{\"o}ther, Maximilian  and
      Canova, Niklas  and
      Challier, Camille  and
      Charmillot, Cl{\'e}ment  and
      Coles, Jonathan  and
      Deriu, Jan Milan  and
      Devos, Arnout  and
      Drescher, Lukas  and
      Dzenhaliou, Daniil  and
      Ehrmann, Maud  and
      Fan, Dongyang  and
      Fan, Simin  and
      Gao, Silin  and
      Gila, Miguel  and
      Grandury, Mar{\'i}a  and
      Hashemi, Diba  and
      Hoyle, Alexander Miserlis  and
      Jiang, Jiaming  and
      Klein, Mark  and
      Kucharavy, Andrei  and
      Kucherenko, Anastasiia  and
      L{\"u}beck, Frederike  and
      Machacek, Roman  and
      Manitaras, Theofilos Ioannis  and
      Marfurt, Andreas  and
      Matoba, Kyle  and
      Matrenok, Simon  and
      Mendon{\c{c}}a, Henrique  and
      Mohamed, Fawzi Roberto  and
      Montariol, Syrielle  and
      Mouchel, Luca  and
      Najem-Meyer, Sven  and
      Ni, Jingwei  and
      Oliva, Gennaro  and
      Pagliardini, Matteo  and
      Palme, Elia  and
      Panferov, Andrei  and
      Paoletti, L{\'e}o  and
      Passerini, Marco  and
      Pavlov, Ivan  and
      Poiroux, Auguste  and
      Ponkshe, Kaustubh  and
      Ranchin, Nathan  and
      Rando, Javier  and
      Sauser, Mathieu  and
      Saydaliev, Jakhongir  and
      Sayfiddinov, Mukhammadali  and
      Schneider, Marian  and
      Schuppli, Stefano  and
      Scialanga, Marco  and
      Semenov, Andrei  and
      Shridhar, Kumar  and
      Singhal, Raghav  and
      Sotnikova, Anna  and
      Sternfeld, Alexander  and
      Tarun, Ayush Kumar  and
      Teiletche, Paul  and
      Vamvas, Jannis  and
      Yao, Xiaozhe  and
      Zhao, Hao  and
      Ilic, Alexander  and
      Klimovic, Ana  and
      Krause, Andreas  and
      Gulcehre, Caglar  and
      Rosenthal, David  and
      Ash, Elliott  and
      Tram{\`e}r, Florian  and
      VandeVondele, Joost  and
      Veraldi, Livio  and
      Rajman, Martin  and
      Schulthess, Thomas C.  and
      Hoefler, Torsten  and
      Bosselut, Antoine  and
      Jaggi, Martin  and
      Schlag, Imanol},
    editor = "Liakata, Maria  and
      Moreira, Viviane P.  and
      Zhang, Jiajun  and
      Jurgens, David",
    booktitle = "Proceedings of the 64th Annual Meeting of the {A}ssociation for {C}omputational {L}inguistics (Volume 1: Long Papers)",
    month = jul,
    year = "2026",
    address = "San Diego, California, United States",
    publisher = "Association for Computational Linguistics",
    url = "https://aclanthology.org/2026.acl-long.2172/",
    doi = "10.18653/v1/2026.acl-long.2172",
    pages = "46877--46955",
    ISBN = "979-8-89176-390-6"
}

@inproceedings{
    lightman2024lets,
    title={Let's Verify Step by Step},
    author={Hunter Lightman and Vineet Kosaraju and Yuri Burda and Harrison Edwards and Bowen Baker and Teddy Lee and Jan Leike and John Schulman and Ilya Sutskever and Karl Cobbe},
    booktitle={The Twelfth International Conference on Learning Representations},
    year={2024},
    url={https://openreview.net/forum?id=v8L0pN6EOi}
    }

@inproceedings{
    hendrycks2021measuring,
    title={Measuring Mathematical Problem Solving With the {MATH} Dataset},
    author={Dan Hendrycks and Collin Burns and Saurav Kadavath and Akul Arora and Steven Basart and Eric Tang and Dawn Song and Jacob Steinhardt},
    booktitle={Thirty-fifth Conference on Neural Information Processing Systems Datasets and Benchmarks Track (Round 2)},
    year={2021},
    url={https://openreview.net/forum?id=7Bywt2mQsCe}
    }

@inproceedings{he-etal-2024-olympiadbench,
    title = "{O}lympiad{B}ench: A Challenging Benchmark for Promoting {AGI} with Olympiad-Level Bilingual Multimodal Scientific Problems",
    author = "He, Chaoqun  and
      Luo, Renjie  and
      Bai, Yuzhuo  and
      Hu, Shengding  and
      Thai, Zhen  and
      Shen, Junhao  and
      Hu, Jinyi  and
      Han, Xu  and
      Huang, Yujie  and
      Zhang, Yuxiang  and
      Liu, Jie  and
      Qi, Lei  and
      Liu, Zhiyuan  and
      Sun, Maosong",
    editor = "Ku, Lun-Wei  and
      Martins, Andre  and
      Srikumar, Vivek",
    booktitle = "Proceedings of the 62nd Annual Meeting of the Association for Computational Linguistics (Volume 1: Long Papers)",
    month = aug,
    year = "2024",
    address = "Bangkok, Thailand",
    publisher = "Association for Computational Linguistics",
    url = "https://aclanthology.org/2024.acl-long.211/",
    doi = "10.18653/v1/2024.acl-long.211",
    pages = "3828--3850"
}

@misc{lighteval,
  author = {Habib, Nathan and Fourrier, Clémentine and Kydlíček, Hynek and Wolf, Thomas and Tunstall, Lewis},
  title = {LightEval: A lightweight framework for LLM evaluation},
  year = {2023},
  version = {0.11.0},
  url = {https://github.com/huggingface/lighteval}
}

@article{Chen2021evaluating,
  author       = {Mark Chen and
                  Jerry Tworek and
                  Heewoo Jun and
                  Qiming Yuan and
                  Henrique Pond{\'{e}} de Oliveira Pinto and
                  Jared Kaplan and
                  Harri Edwards and
                  Yuri Burda and
                  Nicholas Joseph and
                  Greg Brockman and
                  Alex Ray and
                  Raul Puri and
                  Gretchen Krueger and
                  Michael Petrov and
                  Heidy Khlaaf and
                  Girish Sastry and
                  Pamela Mishkin and
                  Brooke Chan and
                  Scott Gray and
                  Nick Ryder and
                  Mikhail Pavlov and
                  Alethea Power and
                  Lukasz Kaiser and
                  Mohammad Bavarian and
                  Clemens Winter and
                  Philippe Tillet and
                  Felipe Petroski Such and
                  Dave Cummings and
                  Matthias Plappert and
                  Fotios Chantzis and
                  Elizabeth Barnes and
                  Ariel Herbert{-}Voss and
                  William Hebgen Guss and
                  Alex Nichol and
                  Alex Paino and
                  Nikolas Tezak and
                  Jie Tang and
                  Igor Babuschkin and
                  Suchir Balaji and
                  Shantanu Jain and
                  William Saunders and
                  Christopher Hesse and
                  Andrew N. Carr and
                  Jan Leike and
                  Joshua Achiam and
                  Vedant Misra and
                  Evan Morikawa and
                  Alec Radford and
                  Matthew Knight and
                  Miles Brundage and
                  Mira Murati and
                  Katie Mayer and
                  Peter Welinder and
                  Bob McGrew and
                  Dario Amodei and
                  Sam McCandlish and
                  Ilya Sutskever and
                  Wojciech Zaremba},
  title        = {Evaluating Large Language Models Trained on Code},
  journal      = {CoRR},
  volume       = {abs/2107.03374},
  year         = {2021},
  url          = {https://arxiv.org/abs/2107.03374},
  eprinttype   = {arXiv},
  eprint       = {2107.03374},
  bibsource    = {dblp computer science bibliography, https://dblp.org}
}

@inproceedings{
    xu2025magpie,
    title={Magpie: Alignment Data Synthesis from Scratch by Prompting Aligned {LLM}s with Nothing},
    author={Zhangchen Xu and Fengqing Jiang and Luyao Niu and Yuntian Deng and Radha Poovendran and Yejin Choi and Bill Yuchen Lin},
    booktitle={The Thirteenth International Conference on Learning Representations},
    year={2025},
    url={https://openreview.net/forum?id=Pnk7vMbznK}
    }

@inproceedings{
    jiang2024wildteaming,
    title={WildTeaming at Scale: From In-the-Wild Jailbreaks to (Adversarially) Safer Language Models},
    author={Liwei Jiang and Kavel Rao and Seungju Han and Allyson Ettinger and Faeze Brahman and Sachin Kumar and Niloofar Mireshghallah and Ximing Lu and Maarten Sap and Yejin Choi and Nouha Dziri},
    booktitle={The Thirty-eighth Annual Conference on Neural Information Processing Systems},
    year={2024},
    url={https://openreview.net/forum?id=n5R6TvBVcX}
    }

@inproceedings{han2024WildGuard,
    author = {Han, Seungju and Rao, Kavel and Ettinger, Allyson and Jiang, Liwei and Lin, Bill Yuchen and Lambert, Nathan and Choi, Yejin and Dziri, Nouha},
    title = {WILDGUARD: open one-stop moderation tools for safety risks, jailbreaks, and refusals of LLMs},
    year = {2024},
    isbn = {9798331314385},
    publisher = {Curran Associates Inc.},
    address = {Red Hook, NY, USA},
    booktitle = {Proceedings of the 38th International Conference on Neural Information Processing Systems},
    articleno = {261},
    numpages = {39},
    location = {Vancouver, BC, Canada},
    series = {NIPS '24}
    }

@inproceedings{Huang2024TrustLLM,
    author = {Huang, Yue and Sun, Lichao and Wang, Haoran and Wu, Siyuan and Zhang, Qihui and Li, Yuan and Gao, Chujie and Huang, Yixin and Lyu, Wenhan and Zhang, Yixuan and Li, Xiner and Sun, Hanchi and Liu, Zhengliang and Liu, Yixin and Wang, Yijue and Zhang, Zhikun and Vidgen, Bertie and Kailkhura, Bhavya and Xiong, Caiming and Xiao, Chaowei and Li, Chunyuan and Xing, Eric and Huang, Furong and Liu, Hao and Ji, Heng and Wang, Hongyi and Zhang, Huan and Yao, Huaxiu and Kellis, Manolis and Zitnik, Marinka and Jiang, Meng and Bansal, Mohit and Zou, James and Pei, Jian and Liu, Jian and Gao, Jianfeng and Han, Jiawei and Zhao, Jieyu and Tang, Jiliang and Wang, Jindong and Vanschoren, Joaquin and Mitchell, John C and Shu, Kai and Xu, Kaidi and Chang, Kai-Wei and He, Lifang and Huang, Lifu and Backes, Michael and Gong, Neil Zhenqiang and Yu, Philip S. and Chen, Pin-Yu and Gu, Quanquan and Xu, Ran and Ying, Rex and Ji, Shuiwang and Jana, Suman and Chen, Tianlong and Liu, Tianming and Zhou, Tianyi and Wang, William and Li, Xiang and Zhang, Xiangliang and Wang, Xiao and Xie, Xing and Chen, Xun and Wang, Xuyu and Liu, Yan and Ye, Yanfang and Cao, Yinzhi and Chen, Yong and Zhao, Yue},
    title = {Position: TRUSTLLM: trustworthiness in large language models},
    year = {2024},
    publisher = {JMLR.org},
    booktitle = {Proceedings of the 41st International Conference on Machine Learning},
    articleno = {813},
    numpages = {105},
    location = {Vienna, Austria},
    series = {ICML'24}
    }

@inproceedings{rottger-etal-2024-xstest,
    title = "{XST}est: A Test Suite for Identifying Exaggerated Safety Behaviours in Large Language Models",
    author = {R{\"o}ttger, Paul  and
      Kirk, Hannah  and
      Vidgen, Bertie  and
      Attanasio, Giuseppe  and
      Bianchi, Federico  and
      Hovy, Dirk},
    editor = "Duh, Kevin  and
      Gomez, Helena  and
      Bethard, Steven",
    booktitle = "Proceedings of the 2024 Conference of the North American Chapter of the Association for Computational Linguistics: Human Language Technologies (Volume 1: Long Papers)",
    month = jun,
    year = "2024",
    address = "Mexico City, Mexico",
    publisher = "Association for Computational Linguistics",
    url = "https://aclanthology.org/2024.naacl-long.301/",
    doi = "10.18653/v1/2024.naacl-long.301",
    pages = "5377--5400"
}

@misc{jegham2025hungryaibenchmarkingenergy,
      title={How Hungry is AI? Benchmarking Energy, Water, and Carbon Footprint of LLM Inference}, 
      author={Nidhal Jegham and Marwan Abdelatti and Chan Young Koh and Lassad Elmoubarki and Abdeltawab Hendawi},
      year={2025},
      eprint={2505.09598},
      archivePrefix={arXiv},
      primaryClass={cs.CY},
      url={https://arxiv.org/abs/2505.09598}, 
}

@inproceedings{Luccioni2024Power,
    author = {Luccioni, Sasha and Jernite, Yacine and Strubell, Emma},
    title = {Power Hungry Processing: Watts Driving the Cost of AI Deployment?},
    year = {2024},
    isbn = {9798400704505},
    publisher = {Association for Computing Machinery},
    address = {New York, NY, USA},
    url = {https://doi.org/10.1145/3630106.3658542},
    doi = {10.1145/3630106.3658542},
    booktitle = {Proceedings of the 2024 ACM Conference on Fairness, Accountability, and Transparency},
    pages = {85–99},
    numpages = {15},
    location = {Rio de Janeiro, Brazil},
    series = {FAccT '24}
    }

@inproceedings{Luccioni2025Jevons,
    author = {Luccioni, Alexandra Sasha and Strubell, Emma and Crawford, Kate},
    title = {From Efficiency Gains to Rebound Effects: The Problem of Jevons' Paradox in AI's Polarized Environmental Debate},
    year = {2025},
    isbn = {9798400714825},
    publisher = {Association for Computing Machinery},
    address = {New York, NY, USA},
    url = {https://doi.org/10.1145/3715275.3732007},
    doi = {10.1145/3715275.3732007},
    booktitle = {Proceedings of the 2025 ACM Conference on Fairness, Accountability, and Transparency},
    pages = {76–88},
    numpages = {13},
    location = {
    },
    series = {FAccT '25}
    }

@inproceedings{
    morrison2025holistically,
    title={Holistically Evaluating the Environmental Impact of Creating Language Models},
    author={Jacob Morrison and Clara Na and Jared Fernandez and Tim Dettmers and Emma Strubell and Jesse Dodge},
    booktitle={The Thirteenth International Conference on Learning Representations},
    year={2025},
    url={https://openreview.net/forum?id=04qx93Viwj}
    }

@article{JMLR:v24:23-0569,
  author  = {Peter Henderson and Xuechen Li and Dan Jurafsky and Tatsunori Hashimoto and Mark A. Lemley and Percy Liang},
  title   = {Foundation Models and Fair Use},
  journal = {Journal of Machine Learning Research},
  year    = {2023},
  volume  = {24},
  number  = {400},
  pages   = {1--79},
  url     = {http://jmlr.org/papers/v24/23-0569.html}
}

@misc{lozhkov2025openr1math220k,
      title={OpenR1-Math-220k},
      author={Anton Lozhkov and Hynek Kydlíček and Loubna Ben Allal and Guilherme Penedo and Edward Beeching and Quentin Gallouédec and Nathan Habib and Lewis Tunstall and Leandro von Werra},
      year={2025},
      publisher = {Hugging Face},
      journal = {Hugging Face repository},
      howpublished = {\url{https://huggingface.co/datasets/open-r1/OpenR1-Math-220k}}
}

@misc{
    yang2024shadow,
    title={Shadow Alignment: The Ease of Subverting Safely-Aligned Language Models},
    author={Xianjun Yang and Xiao Wang and Qi Zhang and Linda Ruth Petzold and William Yang Wang and Xun Zhao and Dahua Lin},
    year={2024},
    url={https://openreview.net/forum?id=rg0vQmkB7F}
    }

@inproceedings{zhan-etal-2024-removing,
    title = "Removing {RLHF} Protections in {GPT}-4 via Fine-Tuning",
    author = "Zhan, Qiusi  and
      Fang, Richard  and
      Bindu, Rohan  and
      Gupta, Akul  and
      Hashimoto, Tatsunori  and
      Kang, Daniel",
    editor = "Duh, Kevin  and
      Gomez, Helena  and
      Bethard, Steven",
    booktitle = "Proceedings of the 2024 Conference of the North American Chapter of the Association for Computational Linguistics: Human Language Technologies (Volume 2: Short Papers)",
    month = jun,
    year = "2024",
    address = "Mexico City, Mexico",
    publisher = "Association for Computational Linguistics",
    url = "https://aclanthology.org/2024.naacl-short.59/",
    doi = "10.18653/v1/2024.naacl-short.59",
    pages = "681--687"
}

@inproceedings{
    kim2025rethinking,
    title={Rethinking Safety in {LLM} Fine-tuning: An Optimization Perspective},
    author={Minseon Kim and Jin Myung Kwak and Lama Alssum and Bernard Ghanem and Philip Torr and David Krueger and Fazl Barez and Adel Bibi},
    booktitle={Second Conference on Language Modeling},
    year={2025},
    url={https://openreview.net/forum?id=ZnOoEA2nDn}
    }

@misc{safety-eval,
  author = {AllenAI},
  title = {Ai2 Safety Tool (Evaluation Suite)},
  year = {2024},
  url = {https://github.com/allenai/safety-eval}
}

@book{collett_2025, 
    title={The Impact of Generative AI on the Novel}, 
    publisher={Minderoo Centre for Technology and Democracy Cambridge},
    url={https://www.repository.cam.ac.uk/handle/1810/391291}, 
    DOI={10.17863/CAM.122470}, 
    institution={Minderoo Centre for Technology and Democracy}, 
    author={Collett, Clementine}, 
    year={2025} 
}

@techreport{mammen2024a,
  institution = {University of Oxford},
  edition = {},
  number = {},
  journal = {},
  booktitle = {},
  pages = {},
  publisher = {University of Oxford},
  school = {},
  title = {Creativity, artificial intelligence, and the requirement of human authors and inventors in copyright and patent law},
  volume = {},
  author = {Mammen, CE and Collyer, M and Dolin, R and Gangjee, DS and Melham, T and Mustaklem, M and Sundaralingam, P and Wang-Maścianica, V},
  editor = {},
  year = {2024},
  series = {}
}

@misc{vonwerra2020trl,
  title   = {{TRL: Transformers Reinforcement Learning}},
  author  = {von Werra, Leandro and Belkada, Younes and Tunstall, Lewis and Beeching, Edward and Thrush, Tristan and Lambert, Nathan and Huang, Shengyi and Rasul, Kashif and Gallouédec, Quentin},
  license = {Apache-2.0},
  url     = {https://github.com/huggingface/trl},
  year    = {2020}
}

@misc{rajbhandari2020zero,
    author = {Rajbhandari, Samyam and Rasley, Jeff and Ruwase, Olatunji and He, Yuxiong},
    title = {ZeRO: Memory Optimizations Toward Training Trillion Parameter Models},
    howpublished = {ArXiv},
    year = {2020},
    month = {May},
    url = {https://www.microsoft.com/en-us/research/publication/zero-memory-optimizations-toward-training-trillion-parameter-models/},
    }

@misc{wandb,
    title = {Experiment Tracking with Weights and Biases},
    year = {2020},
    note = {Software available from wandb.com},
    url={https://www.wandb.com/},
    author = {Biewald, Lukas},
    }

@article{
    casper2026open,
    title={Open Technical Problems in Open-Weight {AI} Model Risk Management},
    author={Stephen Casper and Kyle O'Brien and Shayne Longpre and Elizabeth Seger and Kevin Klyman and Rishi Bommasani and Aniruddha Nrusimha and Ilia Shumailov and S{\"o}ren Mindermann and Steven Basart and Frank Rudzicz and Kellin Pelrine and Avijit Ghosh and Andrew Strait and Robert Kirk and Dan Hendrycks and Peter Henderson and J Zico Kolter and Geoffrey Irving and Yarin Gal and Yoshua Bengio and Dylan Hadfield-Menell},
    journal={Transactions on Machine Learning Research},
    issn={2835-8856},
    year={2026},
    url={https://openreview.net/forum?id=8QyGLnFkzc},
    note={Survey Certification}
    }

@misc{openr1,
    title = {Open R1: A fully open reproduction of DeepSeek-R1},
    url = {https://github.com/huggingface/open-r1},
    author = {{Hugging Face}},
    month = {January},
    year = {2025}
}

@misc{gemmateam2024gemma2improvingopen,
      title={Gemma 2: Improving Open Language Models at a Practical Size}, 
      author={Gemma-Team and Morgane Riviere and Shreya Pathak and Pier Giuseppe Sessa and Cassidy Hardin and Surya Bhupatiraju and Léonard Hussenot and Thomas Mesnard and Bobak Shahriari and Alexandre Ramé and Johan Ferret and Peter Liu and Pouya Tafti and Abe Friesen and Michelle Casbon and Sabela Ramos and Ravin Kumar and Charline Le Lan and Sammy Jerome and Anton Tsitsulin and Nino Vieillard and Piotr Stanczyk and Sertan Girgin and Nikola Momchev and Matt Hoffman and Shantanu Thakoor and Jean-Bastien Grill and Behnam Neyshabur and Olivier Bachem and Alanna Walton and Aliaksei Severyn and Alicia Parrish and Aliya Ahmad and Allen Hutchison and Alvin Abdagic and Amanda Carl and Amy Shen and Andy Brock and Andy Coenen and Anthony Laforge and Antonia Paterson and Ben Bastian and Bilal Piot and Bo Wu and Brandon Royal and Charlie Chen and Chintu Kumar and Chris Perry and Chris Welty and Christopher A. Choquette-Choo and Danila Sinopalnikov and David Weinberger and Dimple Vijaykumar and Dominika Rogozińska and Dustin Herbison and Elisa Bandy and Emma Wang and Eric Noland and Erica Moreira and Evan Senter and Evgenii Eltyshev and Francesco Visin and Gabriel Rasskin and Gary Wei and Glenn Cameron and Gus Martins and Hadi Hashemi and Hanna Klimczak-Plucińska and Harleen Batra and Harsh Dhand and Ivan Nardini and Jacinda Mein and Jack Zhou and James Svensson and Jeff Stanway and Jetha Chan and Jin Peng Zhou and Joana Carrasqueira and Joana Iljazi and Jocelyn Becker and Joe Fernandez and Joost van Amersfoort and Josh Gordon and Josh Lipschultz and Josh Newlan and Ju-yeong Ji and Kareem Mohamed and Kartikeya Badola and Kat Black and Katie Millican and Keelin McDonell and Kelvin Nguyen and Kiranbir Sodhia and Kish Greene and Lars Lowe Sjoesund and Lauren Usui and Laurent Sifre and Lena Heuermann and Leticia Lago and Lilly McNealus and Livio Baldini Soares and Logan Kilpatrick and Lucas Dixon and Luciano Martins and Machel Reid and Manvinder Singh and Mark Iverson and Martin Görner and Mat Velloso and Mateo Wirth and Matt Davidow and Matt Miller and Matthew Rahtz and Matthew Watson and Meg Risdal and Mehran Kazemi and Michael Moynihan and Ming Zhang and Minsuk Kahng and Minwoo Park and Mofi Rahman and Mohit Khatwani and Natalie Dao and Nenshad Bardoliwalla and Nesh Devanathan and Neta Dumai and Nilay Chauhan and Oscar Wahltinez and Pankil Botarda and Parker Barnes and Paul Barham and Paul Michel and Pengchong Jin and Petko Georgiev and Phil Culliton and Pradeep Kuppala and Ramona Comanescu and Ramona Merhej and Reena Jana and Reza Ardeshir Rokni and Rishabh Agarwal and Ryan Mullins and Samaneh Saadat and Sara Mc Carthy and Sarah Cogan and Sarah Perrin and Sébastien M. R. Arnold and Sebastian Krause and Shengyang Dai and Shruti Garg and Shruti Sheth and Sue Ronstrom and Susan Chan and Timothy Jordan and Ting Yu and Tom Eccles and Tom Hennigan and Tomas Kocisky and Tulsee Doshi and Vihan Jain and Vikas Yadav and Vilobh Meshram and Vishal Dharmadhikari and Warren Barkley and Wei Wei and Wenming Ye and Woohyun Han and Woosuk Kwon and Xiang Xu and Zhe Shen and Zhitao Gong and Zichuan Wei and Victor Cotruta and Phoebe Kirk and Anand Rao and Minh Giang and Ludovic Peran and Tris Warkentin and Eli Collins and Joelle Barral and Zoubin Ghahramani and Raia Hadsell and D. Sculley and Jeanine Banks and Anca Dragan and Slav Petrov and Oriol Vinyals and Jeff Dean and Demis Hassabis and Koray Kavukcuoglu and Clement Farabet and Elena Buchatskaya and Sebastian Borgeaud and Noah Fiedel and Armand Joulin and Kathleen Kenealy and Robert Dadashi and Alek Andreev},
      year={2024},
      eprint={2408.00118},
      archivePrefix={arXiv},
      primaryClass={cs.CL},
      url={https://arxiv.org/abs/2408.00118}, 
}

@misc{grattafiori2024llama3herdmodels,
      title={The Llama 3 Herd of Models}, 
      author={Aaron Grattafiori and Abhimanyu Dubey and Abhinav Jauhri and Abhinav Pandey and Abhishek Kadian and Ahmad Al-Dahle and Aiesha Letman and Akhil Mathur and Alan Schelten and Alex Vaughan and Amy Yang and Angela Fan and Anirudh Goyal and Anthony Hartshorn and Aobo Yang and Archi Mitra and Archie Sravankumar and Artem Korenev and Arthur Hinsvark and Arun Rao and Aston Zhang and Aurelien Rodriguez and Austen Gregerson and Ava Spataru and Baptiste Roziere and Bethany Biron and Binh Tang and Bobbie Chern and Charlotte Caucheteux and Chaya Nayak and Chloe Bi and Chris Marra and Chris McConnell and Christian Keller and Christophe Touret and Chunyang Wu and Corinne Wong and Cristian Canton Ferrer and Cyrus Nikolaidis and Damien Allonsius and Daniel Song and Danielle Pintz and Danny Livshits and Danny Wyatt and David Esiobu and Dhruv Choudhary and Dhruv Mahajan and Diego Garcia-Olano and Diego Perino and Dieuwke Hupkes and Egor Lakomkin and Ehab AlBadawy and Elina Lobanova and Emily Dinan and Eric Michael Smith and Filip Radenovic and Francisco Guzmán and Frank Zhang and Gabriel Synnaeve and Gabrielle Lee and Georgia Lewis Anderson and Govind Thattai and Graeme Nail and Gregoire Mialon and Guan Pang and Guillem Cucurell and Hailey Nguyen and Hannah Korevaar and Hu Xu and Hugo Touvron and Iliyan Zarov and Imanol Arrieta Ibarra and Isabel Kloumann and Ishan Misra and Ivan Evtimov and Jack Zhang and Jade Copet and Jaewon Lee and Jan Geffert and Jana Vranes and Jason Park and Jay Mahadeokar and Jeet Shah and Jelmer van der Linde and Jennifer Billock and Jenny Hong and Jenya Lee and Jeremy Fu and Jianfeng Chi and Jianyu Huang and Jiawen Liu and Jie Wang and Jiecao Yu and Joanna Bitton and Joe Spisak and Jongsoo Park and Joseph Rocca and Joshua Johnstun and Joshua Saxe and Junteng Jia and Kalyan Vasuden Alwala and Karthik Prasad and Kartikeya Upasani and Kate Plawiak and Ke Li and Kenneth Heafield and Kevin Stone and Khalid El-Arini and Krithika Iyer and Kshitiz Malik and Kuenley Chiu and Kunal Bhalla and Kushal Lakhotia and Lauren Rantala-Yeary and Laurens van der Maaten and Lawrence Chen and Liang Tan and Liz Jenkins and Louis Martin and Lovish Madaan and Lubo Malo and Lukas Blecher and Lukas Landzaat and Luke de Oliveira and Madeline Muzzi and Mahesh Pasupuleti and Mannat Singh and Manohar Paluri and Marcin Kardas and Maria Tsimpoukelli and Mathew Oldham and Mathieu Rita and Maya Pavlova and Melanie Kambadur and Mike Lewis and Min Si and Mitesh Kumar Singh and Mona Hassan and Naman Goyal and Narjes Torabi and Nikolay Bashlykov and Nikolay Bogoychev and Niladri Chatterji and Ning Zhang and Olivier Duchenne and Onur Çelebi and Patrick Alrassy and Pengchuan Zhang and Pengwei Li and Petar Vasic and Peter Weng and Prajjwal Bhargava and Pratik Dubal and Praveen Krishnan and Punit Singh Koura and Puxin Xu and Qing He and Qingxiao Dong and Ragavan Srinivasan and Raj Ganapathy and Ramon Calderer and Ricardo Silveira Cabral and Robert Stojnic and Roberta Raileanu and Rohan Maheswari and Rohit Girdhar and Rohit Patel and Romain Sauvestre and Ronnie Polidoro and Roshan Sumbaly and Ross Taylor and Ruan Silva and Rui Hou and Rui Wang and Saghar Hosseini and Sahana Chennabasappa and Sanjay Singh and Sean Bell and Seohyun Sonia Kim and Sergey Edunov and Shaoliang Nie and Sharan Narang and Sharath Raparthy and Sheng Shen and Shengye Wan and Shruti Bhosale and Shun Zhang and Simon Vandenhende and Soumya Batra and Spencer Whitman and Sten Sootla and Stephane Collot and Suchin Gururangan and Sydney Borodinsky and Tamar Herman and Tara Fowler and Tarek Sheasha and Thomas Georgiou and Thomas Scialom and Tobias Speckbacher and Todor Mihaylov and Tong Xiao and Ujjwal Karn and Vedanuj Goswami and Vibhor Gupta and Vignesh Ramanathan and Viktor Kerkez and Vincent Gonguet and Virginie Do and Vish Vogeti and Vítor Albiero and Vladan Petrovic and Weiwei Chu and Wenhan Xiong and Wenyin Fu and Whitney Meers and Xavier Martinet and Xiaodong Wang and Xiaofang Wang and Xiaoqing Ellen Tan and Xide Xia and Xinfeng Xie and Xuchao Jia and Xuewei Wang and Yaelle Goldschlag and Yashesh Gaur and Yasmine Babaei and Yi Wen and Yiwen Song and Yuchen Zhang and Yue Li and Yuning Mao and Zacharie Delpierre Coudert and Zheng Yan and Zhengxing Chen and Zoe Papakipos and Aaditya Singh and Aayushi Srivastava and Abha Jain and Adam Kelsey and Adam Shajnfeld and Adithya Gangidi and Adolfo Victoria and Ahuva Goldstand and Ajay Menon and Ajay Sharma and Alex Boesenberg and Alexei Baevski and Allie Feinstein and Amanda Kallet and Amit Sangani and Amos Teo and Anam Yunus and Andrei Lupu and Andres Alvarado and Andrew Caples and Andrew Gu and Andrew Ho and Andrew Poulton and Andrew Ryan and Ankit Ramchandani and Annie Dong and Annie Franco and Anuj Goyal and Aparajita Saraf and Arkabandhu Chowdhury and Ashley Gabriel and Ashwin Bharambe and Assaf Eisenman and Azadeh Yazdan and Beau James and Ben Maurer and Benjamin Leonhardi and Bernie Huang and Beth Loyd and Beto De Paola and Bhargavi Paranjape and Bing Liu and Bo Wu and Boyu Ni and Braden Hancock and Bram Wasti and Brandon Spence and Brani Stojkovic and Brian Gamido and Britt Montalvo and Carl Parker and Carly Burton and Catalina Mejia and Ce Liu and Changhan Wang and Changkyu Kim and Chao Zhou and Chester Hu and Ching-Hsiang Chu and Chris Cai and Chris Tindal and Christoph Feichtenhofer and Cynthia Gao and Damon Civin and Dana Beaty and Daniel Kreymer and Daniel Li and David Adkins and David Xu and Davide Testuggine and Delia David and Devi Parikh and Diana Liskovich and Didem Foss and Dingkang Wang and Duc Le and Dustin Holland and Edward Dowling and Eissa Jamil and Elaine Montgomery and Eleonora Presani and Emily Hahn and Emily Wood and Eric-Tuan Le and Erik Brinkman and Esteban Arcaute and Evan Dunbar and Evan Smothers and Fei Sun and Felix Kreuk and Feng Tian and Filippos Kokkinos and Firat Ozgenel and Francesco Caggioni and Frank Kanayet and Frank Seide and Gabriela Medina Florez and Gabriella Schwarz and Gada Badeer and Georgia Swee and Gil Halpern and Grant Herman and Grigory Sizov and Guangyi and Zhang and Guna Lakshminarayanan and Hakan Inan and Hamid Shojanazeri and Han Zou and Hannah Wang and Hanwen Zha and Haroun Habeeb and Harrison Rudolph and Helen Suk and Henry Aspegren and Hunter Goldman and Hongyuan Zhan and Ibrahim Damlaj and Igor Molybog and Igor Tufanov and Ilias Leontiadis and Irina-Elena Veliche and Itai Gat and Jake Weissman and James Geboski and James Kohli and Janice Lam and Japhet Asher and Jean-Baptiste Gaya and Jeff Marcus and Jeff Tang and Jennifer Chan and Jenny Zhen and Jeremy Reizenstein and Jeremy Teboul and Jessica Zhong and Jian Jin and Jingyi Yang and Joe Cummings and Jon Carvill and Jon Shepard and Jonathan McPhie and Jonathan Torres and Josh Ginsburg and Junjie Wang and Kai Wu and Kam Hou U and Karan Saxena and Kartikay Khandelwal and Katayoun Zand and Kathy Matosich and Kaushik Veeraraghavan and Kelly Michelena and Keqian Li and Kiran Jagadeesh and Kun Huang and Kunal Chawla and Kyle Huang and Lailin Chen and Lakshya Garg and Lavender A and Leandro Silva and Lee Bell and Lei Zhang and Liangpeng Guo and Licheng Yu and Liron Moshkovich and Luca Wehrstedt and Madian Khabsa and Manav Avalani and Manish Bhatt and Martynas Mankus and Matan Hasson and Matthew Lennie and Matthias Reso and Maxim Groshev and Maxim Naumov and Maya Lathi and Meghan Keneally and Miao Liu and Michael L. Seltzer and Michal Valko and Michelle Restrepo and Mihir Patel and Mik Vyatskov and Mikayel Samvelyan and Mike Clark and Mike Macey and Mike Wang and Miquel Jubert Hermoso and Mo Metanat and Mohammad Rastegari and Munish Bansal and Nandhini Santhanam and Natascha Parks and Natasha White and Navyata Bawa and Nayan Singhal and Nick Egebo and Nicolas Usunier and Nikhil Mehta and Nikolay Pavlovich Laptev and Ning Dong and Norman Cheng and Oleg Chernoguz and Olivia Hart and Omkar Salpekar and Ozlem Kalinli and Parkin Kent and Parth Parekh and Paul Saab and Pavan Balaji and Pedro Rittner and Philip Bontrager and Pierre Roux and Piotr Dollar and Polina Zvyagina and Prashant Ratanchandani and Pritish Yuvraj and Qian Liang and Rachad Alao and Rachel Rodriguez and Rafi Ayub and Raghotham Murthy and Raghu Nayani and Rahul Mitra and Rangaprabhu Parthasarathy and Raymond Li and Rebekkah Hogan and Robin Battey and Rocky Wang and Russ Howes and Ruty Rinott and Sachin Mehta and Sachin Siby and Sai Jayesh Bondu and Samyak Datta and Sara Chugh and Sara Hunt and Sargun Dhillon and Sasha Sidorov and Satadru Pan and Saurabh Mahajan and Saurabh Verma and Seiji Yamamoto and Sharadh Ramaswamy and Shaun Lindsay and Shaun Lindsay and Sheng Feng and Shenghao Lin and Shengxin Cindy Zha and Shishir Patil and Shiva Shankar and Shuqiang Zhang and Shuqiang Zhang and Sinong Wang and Sneha Agarwal and Soji Sajuyigbe and Soumith Chintala and Stephanie Max and Stephen Chen and Steve Kehoe and Steve Satterfield and Sudarshan Govindaprasad and Sumit Gupta and Summer Deng and Sungmin Cho and Sunny Virk and Suraj Subramanian and Sy Choudhury and Sydney Goldman and Tal Remez and Tamar Glaser and Tamara Best and Thilo Koehler and Thomas Robinson and Tianhe Li and Tianjun Zhang and Tim Matthews and Timothy Chou and Tzook Shaked and Varun Vontimitta and Victoria Ajayi and Victoria Montanez and Vijai Mohan and Vinay Satish Kumar and Vishal Mangla and Vlad Ionescu and Vlad Poenaru and Vlad Tiberiu Mihailescu and Vladimir Ivanov and Wei Li and Wenchen Wang and Wenwen Jiang and Wes Bouaziz and Will Constable and Xiaocheng Tang and Xiaojian Wu and Xiaolan Wang and Xilun Wu and Xinbo Gao and Yaniv Kleinman and Yanjun Chen and Ye Hu and Ye Jia and Ye Qi and Yenda Li and Yilin Zhang and Ying Zhang and Yossi Adi and Youngjin Nam and Yu and Wang and Yu Zhao and Yuchen Hao and Yundi Qian and Yunlu Li and Yuzi He and Zach Rait and Zachary DeVito and Zef Rosnbrick and Zhaoduo Wen and Zhenyu Yang and Zhiwei Zhao and Zhiyu Ma},
      year={2024},
      eprint={2407.21783},
      archivePrefix={arXiv},
      primaryClass={cs.AI},
      url={https://arxiv.org/abs/2407.21783}, 
}

@misc{mistralMinistralMinistraux,
	author = {Mistral-team},
	title = {{U}n {M}inistral, des {M}inistraux | {M}istral {A}{I} --- mistral.ai},
	howpublished = {\url{https://mistral.ai/news/ministraux/}},
	year = {2024},
	note = {[Accessed 02-10-2026]},
}

@inproceedings{inproceedings,
    author = {Shen, Xinyue and Chen, Zeyuan and Backes, Michael and Shen, Yun and Zhang, Yang},
    year = {2024},
    month = {12},
    pages = {1671-1685},
    title = {"Do Anything Now": Characterizing and Evaluating In-The-Wild Jailbreak Prompts on Large Language Models},
    doi = {10.1145/3658644.3670388}
    }

@InProceedings{pmlr-v37-martens15,
  title = 	 {Optimizing Neural Networks with Kronecker-factored Approximate Curvature},
  author = 	 {Martens, James and Grosse, Roger},
  booktitle = 	 {Proceedings of the 32nd International Conference on Machine Learning},
  pages = 	 {2408--2417},
  year = 	 {2015},
  editor = 	 {Bach, Francis and Blei, David},
  volume = 	 {37},
  series = 	 {Proceedings of Machine Learning Research},
  address = 	 {Lille, France},
  month = 	 {07--09 Jul},
  publisher =    {PMLR},
  url = 	 {https://proceedings.mlr.press/v37/martens15.html}
}

@inproceedings{
    walsh2025,
    title={2 {OLM}o 2 Furious ({COLM}{\textquoteright}s Version)},
    author={Evan Pete Walsh and Luca Soldaini and Dirk Groeneveld and Kyle Lo and Shane Arora and Akshita Bhagia and Yuling Gu and Shengyi Huang and Matt Jordan and Nathan Lambert and Dustin Schwenk and Oyvind Tafjord and Taira Anderson and David Atkinson and Faeze Brahman and Christopher Clark and Pradeep Dasigi and Nouha Dziri and Allyson Ettinger and Michal Guerquin and David Heineman and Hamish Ivison and Pang Wei Koh and Jiacheng Liu and Saumya Malik and William Merrill and Lester James Validad Miranda and Jacob Morrison and Tyler Murray and Crystal Nam and Jake Poznanski and Valentina Pyatkin and Aman Rangapur and Michael Schmitz and Sam Skjonsberg and David Wadden and Christopher Wilhelm and Michael Wilson and Luke Zettlemoyer and Ali Farhadi and Noah A. Smith and Hannaneh Hajishirzi},
    booktitle={Second Conference on Language Modeling},
    year={2025},
    url={https://openreview.net/forum?id=2ezugTT9kU}
    }

@misc{qwen2.5,
    title = {Qwen2.5: A Party of Foundation Models},
    url = {https://qwenlm.github.io/blog/qwen2.5/},
    author = {Qwen-Team},
    month = {September},
    year = {2024}
}

@inproceedings{wolf-etal-2020-transformers,
    title = "Transformers: State-of-the-Art Natural Language Processing",
    author = "Thomas Wolf and Lysandre Debut and Victor Sanh and Julien Chaumond and Clement Delangue and Anthony Moi and Pierric Cistac and Tim Rault and Rémi Louf and Morgan Funtowicz and Joe Davison and Sam Shleifer and Patrick von Platen and Clara Ma and Yacine Jernite and Julien Plu and Canwen Xu and Teven Le Scao and Sylvain Gugger and Mariama Drame and Quentin Lhoest and Alexander M. Rush",
    booktitle = "Proceedings of the 2020 Conference on Empirical Methods in Natural Language Processing: System Demonstrations",
    month = oct,
    year = "2020",
    address = "Online",
    publisher = "Association for Computational Linguistics",
    url = "https://aclanthology.org/2020.emnlp-demos.6/",
    pages = "38--45"
}

\appendix

\section{Full results tables}
\label{sec:Appendix - Full results tables}

We provide a detailed, per-benchmark breakdown across models, benchmarks and interventions in Table \ref{tab:Full results table}. For maths, M-500 denotes MATH-500 (MIT license) and Olymp denotes OlympiadBench (Apache 2.0). For jailbreak resistance, Trust denotes TrustLLM (JailbreakTrigger), WGT WildGuardTest (adversarial), WJB:H WildJailbreak (harmful), and DAN DoAnythingNow. We use the versions distributed with AllenAI's \texttt{safety-eval}, all released under ODC-BY. For over-refusal avoidance, we use XSTest (CC BY 4.0), and WJB:B denotes WildJailbreak (benign; ODC-BY). We additionally use WildGuardMix (ODC-BY) to construct $\mathcal{D}_{\text{Safety}}$ and $\mathcal{D}_{\text{Util}}$. All datasets are used for research purposes, consistent with their licenses.

\begin{table*}
    \centering
    \small
    \begin{tabular}{l *{10}{c}}
        \toprule
        & \multicolumn{2}{c}{Math} & \multicolumn{5}{c}{Jailbreak resistance} & \multicolumn{3}{c}{Over-refusal avoidance} \\
        \cmidrule(lr){2-3} \cmidrule(lr){4-8} \cmidrule(lr){9-11}
        Method & M-500 & Olymp & Trust & WGT & WJB:H & DAN & \textbf{Avg.} & XSTest & WJB:B & \textbf{Avg.} \\
        \midrule
        \multicolumn{11}{l}{\textit{Llama-3.1-8B-it}} \\
        Before training & 44.9 & 16.1 & 91.0 & 97.0 & 83.8 & 99.0 & \textbf{92.7} & 94.8 & 81.6 & \textbf{88.2} \\
        Low-LR SFT & 55.6 & 23.2 & 95.0 & 69.7 & 8.3 & 15.7 & \textbf{47.2} & 100.0 & 100.0 & \textbf{100.0} \\
        SafeInstr & 52.6 & 23.0 & 95.0 & 100.0 & 98.7 & 94.0 & \textbf{96.9} & 64.0 & 18.0 & \textbf{41.0} \\
        SafeGrad & 54.8 & 21.8 & 81.2 & 84.0 & 39.8 & 74.7 & \textbf{69.9} & 90.4 & 99.2 & \textbf{94.8} \\
        DER & 53.8 & 21.1 & 92.8 & 86.4 & 43.0 & 82.3 & \textbf{76.1} & 89.2 & 98.8 & \textbf{94.0} \\
        EWC & 51.8 & 20.8 & 96.2 & 89.0 & 58.9 & 65.0 & \textbf{77.3} & 95.6 & 97.2 & \textbf{96.4} \\
        SP-EWC-LoRA & 41.9 & 15.5 & 38.5 & 78.0 & 12.6 & 36.7 & \textbf{41.5} & 99.2 & 100.0 & \textbf{99.6} \\
        SP-EWC-ActSVD & 49.2 & 20.7 & 27.7 & 75.1 & 11.5 & 34.7 & \textbf{37.2} & 100.0 & 100.0 & \textbf{100.0} \\
        SP-EWC-cPCA & 51.0 & 19.8 & 93.5 & 93.8 & 82.5 & 98.3 & \textbf{92.0} & 93.6 & 80.4 & \textbf{87.0} \\
        \midrule
        \multicolumn{11}{l}{\textit{OLMo-2-7B-it}} \\
        Before training & 35.5 & 12.8 & 85.2 & 98.8 & 76.4 & 90.7 & \textbf{87.8} & 93.2 & 97.2 & \textbf{95.2} \\
        Low-LR SFT & 47.5 & 20.5 & 76.5 & 95.8 & 63.5 & 70.3 & \textbf{76.6} & 87.2 & 96.8 & \textbf{92.0} \\
        SafeInstr & 44.5 & 21.3 & 99.2 & 99.7 & 88.6 & 99.0 & \textbf{96.7} & 78.4 & 87.2 & \textbf{82.8} \\
        SafeGrad & 33.1 & 11.7 & 88.5 & 98.5 & 77.4 & 93.0 & \textbf{89.4} & 93.6 & 96.8 & \textbf{95.2} \\
        DER & 36.4 & 12.2 & 91.2 & 98.5 & 75.8 & 97.7 & \textbf{90.8} & 93.2 & 94.0 & \textbf{93.6} \\
        EWC & 39.5 & 14.1 & 85.8 & 98.5 & 76.8 & 91.0 & \textbf{88.0} & 92.8 & 94.4 & \textbf{93.6} \\
        SP-EWC-LoRA & 30.6 & 8.3 & 82.8 & 98.2 & 71.8 & 86.0 & \textbf{84.7} & 91.6 & 96.4 & \textbf{94.0} \\
        SP-EWC-ActSVD & 40.9 & 15.9 & 84.2 & 99.4 & 73.8 & 91.7 & \textbf{87.3} & 91.2 & 96.4 & \textbf{93.8} \\
        SP-EWC-cPCA & 43.1 & 16.7 & 85.2 & 98.8 & 74.8 & 89.0 & \textbf{87.0} & 88.8 & 96.0 & \textbf{92.4} \\
        \bottomrule
    \end{tabular}
    \caption{Per-benchmark results for Llama-3.1-8B-it and OLMo-2-7B-it trained on maths with each intervention. All scores are in \%, and higher is better across all benchmarks.}
    \label{tab:Full results table}
\end{table*}

\section{Further details on implementation}
\label{sec:Appendix - Further details on implementation}
\paragraph{Software}All training runs were run on 4xH200 nodes running docker image nvidia/cuda:12.9.1-devel-ubuntu22.04, CUDA version 12.9 and driver version 570.124.06. We run training jobs using TRL \cite{vonwerra2020trl} and DeepSpeed ZeRO stage 1 \cite{rajbhandari2020zero}, forking the \texttt{open-r1} repository by \citet{openr1}, which has an Apache-2.0 license. We track training runs with Weights \& Biases \citep{wandb}.

\paragraph{Data pre-processing}We train models on data from \texttt{open-r1/Mixture-of-Thoughts}\footnote{\url{https://huggingface.co/datasets/open-r1/Mixture-of-Thoughts}}. We train models to generate chains-of-thought on maths problems using the maths subset, i.e. OpenR1-Math-220k \citep{lozhkov2025openr1math220k}, which has an Apache 2.0 license. For each sample, we apply the chat template, and then discard samples larger than 4,096 tokens, with 126k samples remaining. The chat template is edited to request a chain-of-thought with $<$think$>$...$<$/think$>$ tags, and \{\% generation \%\} tags are added via the jinja template such that TRL's \texttt{assistant\_only\_loss} works correctly.

\paragraph{Evaluation}We find \texttt{LightEval}'s default of pass@k:k=1\&n=1 \citep{Chen2021evaluating} to be noisy across evaluation runs, and develop custom implementations with pass@k:k=1\&n=4 to reduce cross-run variance. We evaluate models with a temperature of 0.6, top\_p of 0.95, and max\_tokens of 4096.

\paragraph{Preparing the Fisher and projection matrices}Before training, we compute the relevant statistics on $\theta^{ref}$, processing samples one at a time. For each prompt, we keep the first 20 tokens of a completion, ignoring prompt and chat template tokens. For locating the subspace with cPCA, we draw 10,000 samples from $\mathcal{D}_{\text{Safety}}$ and 10,000 from $\mathcal{D}_{\text{Util}}$, splitting inputs across GPUs to exploit parallelism. Without computing gradients, we cache the inputs to every linear layer in the model. We note that the query, key and value projections take the same input, as do the gate and up projections of the MLP. As such, we can avoid collecting these inputs twice. For each of these recorded inputs $x_t$ for some completion token $t$, we accumulate the sum of outer products $\sum_t x_t x_t^\top$, add them across GPUs and divide by the total number of tokens. This yields a $d_{in} \times d_{in}$ matrix per layer and dataset, $G_{\text{safe}}$ and $G_{\text{util}}$. Note that a linear layer with weights $W$ maps $x$ to $Wx$, and the corresponding matrix for its outputs is $W G W^\top$, which means that we never need to store the outputs themselves. For each layer we then form $W G_{\text{safe}} W^\top - \alpha\, WG_{\text{util}} W^\top$ in FP32. We perform eigendecomposition and keep the $r$ eigenvectors with the largest eigenvalues as the columns of $U \in \mathbb{R}^{d_{out} \times r}$. Connecting to the intuition in Section \ref{sec:From contrastive principal component analysis}, these directions should be the ones with maximal variance in safety activations in contrast to utility activations. 
We then perform a second pass across a new set of 10,000 examples from $\mathcal{D}_{\text{Safety}}$ to collect gradients, i.e. $g \in \mathbb{R}^{d_{out} \times d_{in}}$. We project the gradients onto the subspace with the per-layer projection matrix, yielding $\tilde{g} = U^\top g$. We add $g^2$ (element-wise) to the diagonal Fisher, and add $\tilde{g}\tilde{g}^\top$ to the $r \times r$ subspace Fisher. Both are averaged over examples on each GPU and then across GPUs with all-reduce. We exclude embeddings, normalisation layers and the LM head. We use $r=8$ and $\alpha=100$.

\section{Details on computational overhead}
\label{sec:Appendix - Details on computational overhead}
In this section, we discuss the computational overhead introduced by the various methods. We assume a 4xH200 (141GB) setup as described above, with an effective batch size 16, so the number of optimiser steps is identical across all methods. We report wall-clock training time in Table \ref{tab:compute overhead}, including pre-computation of the Fisher or projection matrices where applicable, as logged by Weights \& Biases.

\begin{itemize}
    \item \textbf{Standard SFT} is the baseline, taking 12.5 hours, as matched by SafeInstr.
    \item \textbf{EWC} requires a pass over the safety data for estimating the diagonal FIM before training ($\sim 10$ minutes). During training, it adds an element-wise penalty per step. It includes storing $\theta^{ref}$ to compute per-parameter differences, as well as the diagonal FIM, both of which require the same amount of memory as the model itself. We keep one copy per GPU. EWC adds $\sim 15\%$ to training time.
    \item \textbf{\method} adds close to no memory or compute overhead. Its addition on top of EWC (accumulating activations for the subspace in the case of ActSVD and cPCA, and then projecting the Fisher) takes 25-28 minutes, which is $\sim 3\%$ of training time, and can be cached and reused. During training, it involves projecting $\Delta$ onto $r$ directions and an $r \times r$ product per layer. As a result, training is just as fast as EWC, with roughly the same peak memory usage.
    \item \textbf{SafeGrad and DER} are both substantially more expensive, because every training step requires additional forward and backward passes on a batch of safety data, on top of a forward pass through $\theta^{ref}$ for the KL term.
\end{itemize}

\begin{table}[h]
    \centering
    \small
    \begin{tabular}{lcc}
        \toprule
        Method & Time (h) & Overhead vs.\ SFT \\
        \midrule
        Low-LR SFT & 12.5 & -- \\
        EWC & 14.3 & +15\% \\
        \method~(cPCA) & 14.2 & +14\% \\
        \method~(ActSVD) & 14.7 & +18\% \\
        \method~(LoRA) & 14.8 & +19\% \\
        SafeGrad & 21.0 & +69\% \\
        DER & 24.1 & +93\% \\
        \bottomrule
    \end{tabular}
    \caption{Wall-clock training time of OLMo-2-7B-it on 4xH200, including pre-computation before training.}
    \label{tab:compute overhead}
\end{table}

\section{Further notes on computing Fisher Information for LLMs}
\label{sec:Appendix - Further notes on computing Fisher Information for LLMs}

We provide further detail on computing the Fisher at LLM scale, drawing from notation and results by \citet{ven2025on} and \citet{li2025fishers}.

Neural networks, such as LLMs, are typically trained via common loss functions such as the negative log-likelihood (such as in Equation \ref{eqn:SFT}). With this loss, we can obtain per-parameter gradients $g_n$ with respect to some data, such as sample $n$ of $\mathcal{D}_{\text{SFT}}$. These gradients are typically a part of model training, but they can also be used for estimating quantities like parameter importance or curvature. Specifically, they can be used to compute the Fisher Information Matrix (FIM), which is a popular way to estimate the sensitivity of the loss with respect to the parameters \citep{elastic_weight_consolidation, li2025fishers}.

In our context, the standard FIM would be defined as

\begin{equation}
\label{eqn:Standard FIM3}
    F^{std}(\theta^{ref};\mathcal{D}_{\text{Safety}}) = \frac{1}{M} \sum_{j=1}^M \mathbb{E}_{\hat{y}_j \sim p(\mathrm{y}|x_j,\theta^{ref})} [\hat{g}_j \hat{g}_j^\top].
\end{equation}

However, this introduces two key issues. The first one is that the standard FIM requires sampling from the reference model's likelihood,  ${\hat{y}_j \sim p(\mathrm{y}|x_j,\theta^{ref})}$, which is expensive. As such, a popular choice is to instead utilise the empirical FIM, which uses the empirical likelihood, i.e. the statistics from the data, $p(y|x_j, \theta^{ref}) \rightarrow \delta(\mathrm{y} - y_j)$. With this, the gradients become

\begin{equation}
    \label{eqn:gradients from data}
    g_j := -\nabla _{\theta^0} \log p(y_j |x_j, \theta^{ref})
\end{equation}

and the empirical FIM is defined as

\begin{equation}
    \label{eqn:Empirical standard FIM}
    \begin{split}
    F^{emp}(\theta^{ref}; \mathcal{D}_{\text{Safety}}) &= \frac{1}{M} \sum_{i=1}^M \mathbb{E}_{\hat{y}_j \sim \delta(\mathrm{y - y_j})} [\hat{g}_j \hat{g}_j^\top]\\
    &= \frac{1}{M} \sum_{j=1}^M g_j g_j^\top.
    \end{split}
\end{equation}

Yet, the second issue remains: computing the empirical FIM includes multiplying the gradient of the entire network by its transpose. Considering the massive scale of LLMs, this operation quickly becomes intractable (for an $8\text{B}$ LLM this is an $8\text{B}^2$-sized object). A common approximation is therefore to only consider the diagonal of the empirical FIM:

\begin{equation}
    \label{eqn:diagonal empirical FIM(appdx)}
    D^{emp} := \mathrm{diag}(F (\theta^{ref};\mathcal{D}_{\text{Safety}})) = \frac{1}{M} \sum_{j=1}^M g_j^2
\end{equation}

\begin{figure*}[h!]

\begin{center}
\includegraphics[width=1\linewidth]{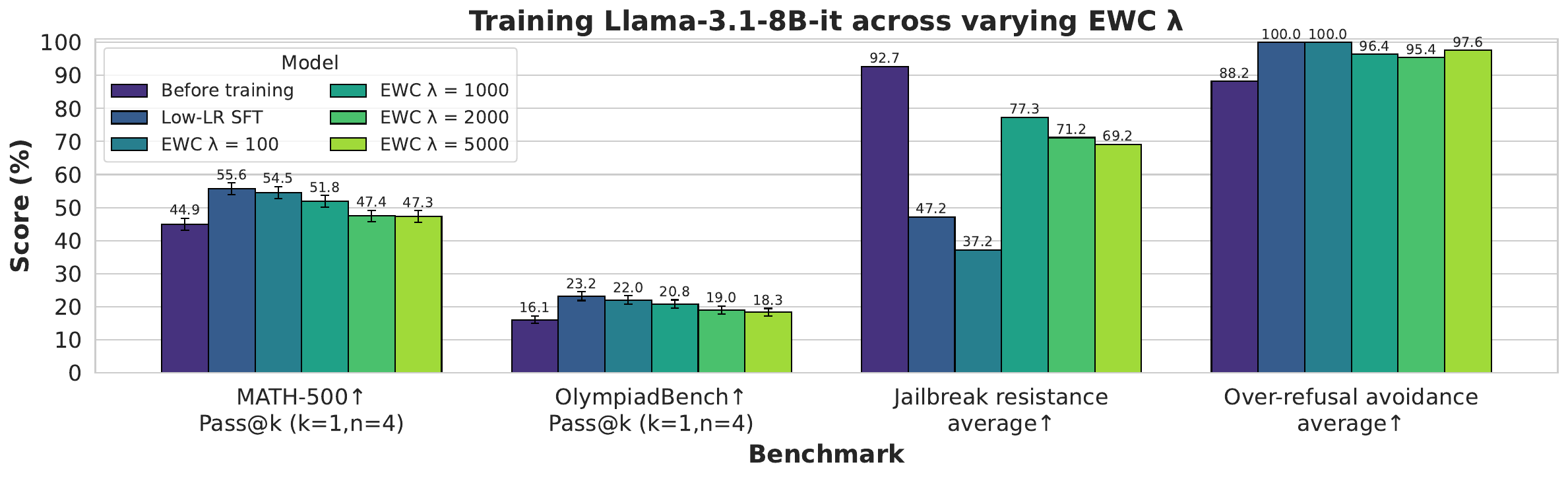}
\end{center}
\caption{The performance of Llama-3.1-8B-it on maths and safety benchmarks when varying the $\lambda$ hyper-parameter which modulates the strength of the EWC penalty.}
\label{fig:Varying the strength of the EWC penalty}
\end{figure*}

In the safety context, this is equivalent to iterating through the entire safety dataset, computing the gradient for each sample, squaring it, and then accumulating it across all samples. This is relatively cheap to do, in that it is similar to standard training except that there are no optimiser or parameter updates. Crucially, it provides an approximation of the sensitivity of the loss on safety data with respect to each parameter. The empirical Fisher has been found to have limitations, such as in the context of natural gradient descent \citep{limitations2019}, and \citet{ven2025on} advocate for more precise approximations.Notably, because we sample $\mathcal{D}_{\text{Safety}}$ directly from $\theta^{ref}$, our approach can be interpreted as a sampled estimate of the true Fisher, as suggested by \citet{ven2025on}.

There are further critical details when computing the empirical Fisher. In standard PyTorch fashion, gradients are computed with respect to batches of data at once. \Citet{ven2025on} shows that this typical batched setup underperforms empirically on MNIST and Split CIFAR-10. We observe similar dynamics even at the scale of 8B LLMs, and are therefore careful to compute our approximation of the Fisher with a batch size of 1. However, in the context of LLMs, the story is complicated further: EWC was originally evaluated on an MLP trained for MNIST, which takes an image as an input and predicts a label as an output. Meanwhile, our context considers language, i.e. sequences of tokens, where the transformer produces predictions for all tokens in the sequence at the same time. Therefore, even just taking the loss on a single sequence with HuggingFace's transformers library \citep{wolf-etal-2020-transformers} arguably differs from the "single input, single label" context. However, our results suggest that our "diagonal of the sequence-level sampled true Fisher" provides a strong baseline in EWC, and a strong method as part of \method~overall.

\begin{figure*}[ht!]
\begin{center}
\includegraphics[width=1\linewidth]{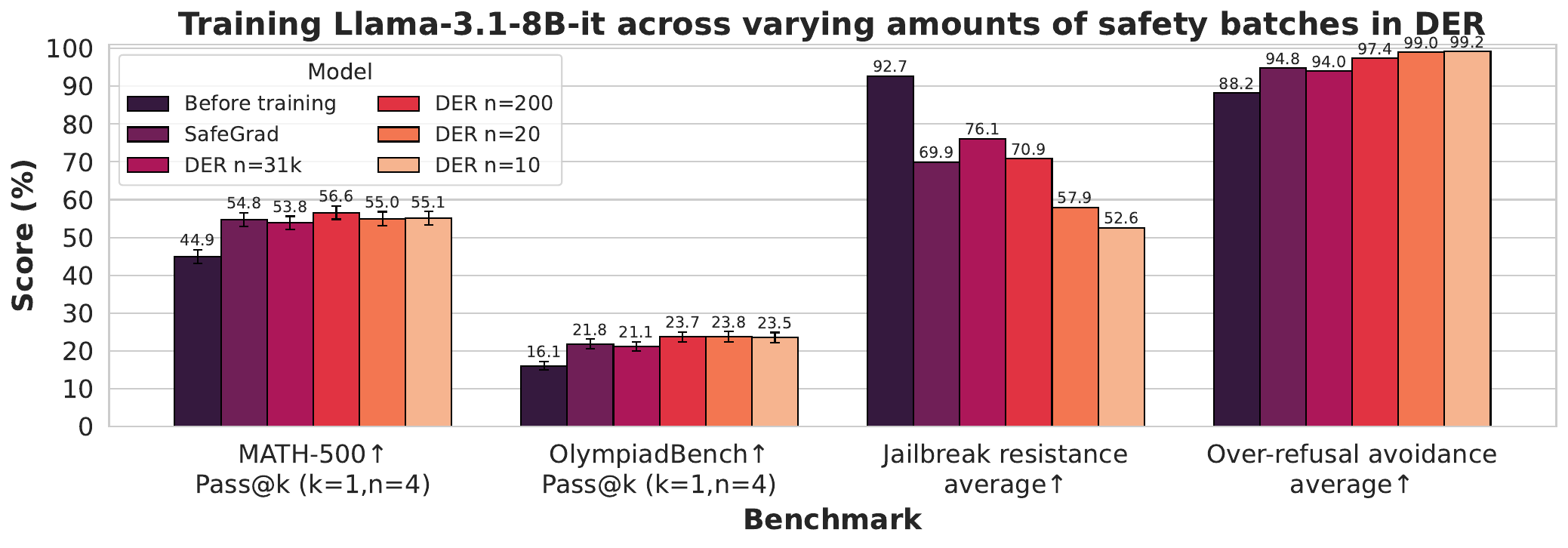}
\end{center}
\caption{The performance of Llama-3.1-8B-it on maths and safety benchmarks when using SafeGrad, or when using DER under varying amounts of safety data.}
\label{fig:SafeGrad and DER under varying amounts of safety data}
\end{figure*}

\section{Varying the strength of the EWC penalty}
\label{sec:Appendix - Varying the strength of the EWC penalty}

We provide results for our EWC penalty sweep in Figure \ref{fig:Varying the strength of the EWC penalty}. We vary $\lambda$, which modulates the strength of the Fisher-weighted parameter difference penalty. We observe that as the value of $\lambda$ increases jailbreak resistance can be maintained better compared to standard SFT. However, for growing values of $\lambda$, the in-domain benefit of fine-tuning continually diminishes, and jailbreak resistance drops again. Intuitively, a higher regulariser strength places more pressure on the model to deviate less from $\theta^{ref}$. As such, this exhibits a pronounced loss in fine-tuning benefit, without maintaining the jailbreak resistance of the reference model faithfully. This is in line with results by \citet{alssum2025unforgottensafetypreservingsafety}, and motivates \method.

\section{Choice of models}
\label{sec:Appendix - Choice of models}
Limiting our scope to models with $10$B parameters or less due to compute constraints, we decide to focus on models that have comparatively stronger safety guardrails, as indicated by jailbreak resistance benchmarks, before training. This is important, as some models demonstrate notable performance on capability benchmarks, but are easily fooled by simple adversarial prompts even without fine-tuning. In that regime, an intervention seeking to prevent safety degradation is harder to motivate, as there is not much to retain in the first place. For this same reason, testing our method on \texttt{base} models makes little sense. Finally, we choose to focus on training non-reasoning models to reason as a testbed for more notable capability and behaviour change.

Table \ref{tab:Choice of models} shows the averages of Gemma-2-9B-it \citep{gemmateam2024gemma2improvingopen}, Llama-3.1-8B-it \citep{grattafiori2024llama3herdmodels}, Ministral-8B-it \citep{mistralMinistralMinistraux}, OLMo-2-7B-it \citep{walsh2025} and Qwen-2.5-7B-it \citep{qwen2.5}. Gemma-2-9B-it is released under the Gemma Terms of Use, Llama-3.1-8B-it under the Llama 3.1 Community License Agreement, Ministral-8B-it under the Mistral AI Research License, and OLMo-2-7B-it and Qwen-2.5-7B-it under Apache 2.0. Our use of all models is for research purposes, consistent with their respective licenses. We find Llama and OLMo to attain considerably better jailbreak resistance scores, and therefore select them for our experiments.

\begin{table}[h]
    \centering
    \small
    \begin{tabular}{lcc}
        \toprule
        Model & JR & OA \\
        \midrule
        Gemma-2-9B-it & 57.8 & 90.8 \\
        Llama-3.1-8B-it & 92.7 & 88.2 \\
        Ministral-8B-it & 29.7 & 97.8 \\
        OLMo-2-7B-it & 87.8 & 95.2 \\
        Qwen-2.5-7B-it & 47.9 & 98.2 \\
        \bottomrule
    \end{tabular}
    \caption{Jailbreak resistance (JR) and over-refusal avoidance (OA) of candidate \texttt{instruct} models before any training, averaged over the same benchmarks as in our main results. All scores in \%, higher is better.}
    \label{tab:Choice of models}
\end{table}

\section{SafeGrad and DER under varying amounts of safety data}
\label{sec:Appendix - SafeGrad and DER under varying amounts of safety data}

We provide results for varying the amount of safety data for DER in Figure \ref{fig:SafeGrad and DER under varying amounts of safety data}, as compared to SafeGrad and the reference model. Our results follow those by \citet{alssum2025unforgottensafetypreservingsafety} in that DER can help mitigate guardrail degradation, though we find it to not fully close the gap. Additionally, we observe that a sufficiently large data distribution for safety batches is critical, with considerably less protection with smaller dataset sizes.

\section{Tuning the learning rate}
\label{sec:Appendix - Varying the learning rate}

We provide results from our learning rate sweep in Table \ref{tab:Varying the learning rate}. In line with findings by \citet{lin2026sft}, lowering the learning rate reduces catastrophic forgetting while still providing most of the in-domain benefit.

\begin{table}[h]
    \centering
    \small
    \begin{tabular}{l *{4}{c}}
        \toprule
        Learning rate & M-500 & Olymp & JR & OA \\
        \midrule
        \multicolumn{5}{l}{\textit{Llama-3.1-8B-it}} \\
        Before training & 44.9 & 16.1 & 92.7 & 88.2 \\
        $1 \times 10^{-6}$ & 55.6 & 23.2 & 47.2 & 100.0 \\
        $5 \times 10^{-6}$ & 55.0 & 23.3 & 23.8 & 99.2 \\
        $1 \times 10^{-5}$ & 54.1 & 24.4 & 17.3 & 99.6 \\
        $5 \times 10^{-5}$ & 42.1 & 16.5 & 24.7 & 90.8 \\
        \midrule
        \multicolumn{5}{l}{\textit{OLMo-2-7B-it}} \\
        Before training & 35.5 & 12.8 & 87.8 & 95.2 \\
        $1 \times 10^{-6}$ & 41.5 & 13.8 & 86.7 & 93.2 \\
        $5 \times 10^{-6}$ & 48.2 & 20.0 & 84.8 & 90.4 \\
        $1 \times 10^{-5}$ & 47.5 & 20.5 & 76.6 & 92.0 \\
        $5 \times 10^{-5}$ & 45.1 & 19.8 & 18.9 & 90.6 \\
        \bottomrule
    \end{tabular}
    \caption{Performance of Llama-3.1-8B-it and OLMo-2-7B-it on maths and safety benchmarks after SFT with different learning rates. All scores in \%, higher is better.}
    \label{tab:Varying the learning rate}
\end{table}

\section{Varying the amount of safety data}
\label{sec:Appendix - Varying the amount of safety data}

\begin{table}[!t]
    \centering
    \small
    \begin{tabular}{l *{4}{c}}
        \toprule
        Safety data & M-500 & Olymp & JR & OA \\
        \midrule
        \multicolumn{5}{l}{\textit{Llama-3.1-8B-it}} \\
        Before training & 44.9 & 16.1 & 92.7 & 88.2 \\
        Low-LR SFT & 55.6 & 23.2 & 47.2 & 100.0 \\
        1\% & 56.1 & 21.4 & 98.7 & 34.2 \\
        2\% & 51.4 & 21.3 & 95.8 & 41.2 \\
        3\% & 52.6 & 23.0 & 96.9 & 41.0 \\
        4\% & 52.5 & 21.3 & 98.2 & 46.2 \\
        5\% & 53.5 & 20.4 & 97.6 & 43.6 \\
        \midrule
        \multicolumn{5}{l}{\textit{OLMo-2-7B-it}} \\
        Before training & 35.5 & 12.8 & 87.8 & 95.2 \\
        Low-LR SFT & 47.5 & 20.5 & 76.6 & 92.0 \\
        1\% & 50.7 & 21.0 & 95.1 & 81.0 \\
        2\% & 51.8 & 21.1 & 95.1 & 81.6 \\
        3\% & 46.5 & 21.9 & 94.6 & 85.8 \\
        4\% & 45.6 & 21.3 & 96.7 & 80.4 \\
        5\% & 44.5 & 21.3 & 96.7 & 82.8 \\
        \bottomrule
    \end{tabular}
    \caption{Performance of Llama-3.1-8B-it and OLMo-2-7B-it on maths and safety benchmarks after mixing varying percentages of safety data into the training data (SafeInstr). We reproduce the finding by \citet{bianchi2024safetytuned} that this maintains alignment very well while causing over-refusal. All scores in \%, higher is better.}
    \label{tab:Varying the amount of safety data}
\end{table}

We provide results from our safety data sweep in Table \ref{tab:Varying the amount of safety data}. We observe that the targeted introduction of safety data into the training process works very well for maintaining jailbreak resistance. However, we also reproduce the finding by \citet{bianchi2024safetytuned} that this causes over-refusal. Finding a balance between these two is a key objective for any safety training method. We find that \method~is able to find a favourable trade-off, while still teaching the new task of reasoning for maths problems. Note that across varying percentages, the number of total samples in the training dataset remains constant, and that the combined dataset is shuffled, which contrasts with the approach proposed by \citet{huang2024Lisa}.

\end{document}